\documentclass[Afour,sageh,times]{sagej}

\usepackage{moreverb,url}

\newcommand\BibTeX{{\rmfamily B\kern-.05em \textsc{i\kern-.025em b}\kern-.08em
T\kern-.1667em\lower.7ex\hbox{E}\kern-.125emX}}

\def\volumeyear{2015}

\usepackage{epsfig}           
\usepackage{mathrsfs}
\usepackage{enumerate}
\usepackage{latexsym}
\usepackage{eufrak}
\usepackage{multirow}
\usepackage{makecell}
\usepackage{array}
\usepackage{graphicx}
\usepackage{epstopdf}

\usepackage{setspace}
\usepackage[linesnumbered,ruled,vlined]{algorithm2e}
\usepackage{natbib}
\usepackage{color}
\usepackage{amsmath}
\usepackage{amsfonts}
\usepackage{amssymb}
\usepackage{bm}

\newcommand{\Eq}[1]  {Equation\ \ref{eq:#1}}

\newcommand{\Fig}[1] {Figure\ \ref{fig:#1}}

\newcommand{\Tbl}[1]  {Table \ref{tbl:#1}}

\newcommand{\Alg}[1] {Algorithm \ref{alg:#1}}

\renewcommand \paragraph[1] {\vspace{0.05cm} \textbf{#1}}

\def\etal{\emph{et al.}\xspace}

\newcommand{\argmin}{\operatornamewithlimits{argmin}}

\usepackage{color}

\begin{document}

\runninghead{Li, et al.}

\title{Model-Driven Feed-Forward Prediction for Manipulation of Deformable Objects}

\author{Yinxiao Li\affilnum{1}, Yan Wang\affilnum{2}, Yonghao Yue\affilnum{1}, Danfei Xu\affilnum{1}, Michael Case\affilnum{1}, Shih-Fu Chang\affilnum{1}\affilnum{2}, Eitan Grinspun\affilnum{1}, and Peter Allen\affilnum{1}}

\affiliation{\affilnum{1}Department of Computer Science, Columbia University\\
\affilnum{2}Department of Electrical Engineering, Columbia University}

\corrauth{Yinxiao Li, Columbia University, Room 450 Mudd 500 W. 120th Street, M.C. 0401
New York, New York 10027}

\email{yli@cs.columbia.edu}


\begin{abstract}

Robotic manipulation of deformable objects is a difficult problem especially because of the complexity of the many different ways  an object can deform. 
Searching such a high dimensional state space  makes it difficult to recognize, track, and manipulate deformable objects.
In this paper, we introduce a \emph{predictive, model-driven} approach to address this challenge, using a pre-computed, simulated database of deformable object models.
Mesh models of common deformable garments are simulated with the garments picked up in multiple different poses under gravity, and stored in a database for fast and efficient retrieval.  To validate this approach, we developed a comprehensive pipeline for manipulating clothing as in a typical laundry task. First, the database is used for category and pose estimation for a garment in an arbitrary position. A fully featured 3D model of the garment is constructed in real-time and volumetric features are then used to obtain the most similar model in the database to predict the object category and pose.
Second, the database can significantly benefit the manipulation of deformable objects via non-rigid registration, providing accurate correspondences between the reconstructed object model and the database models.
Third, the accurate model simulation can also be used to optimize the trajectories for manipulation of deformable objects, such as the folding of garments. Extensive experimental results are shown for the tasks above using a variety of different clothing.

\end{abstract}
\maketitle

\section{Introduction}
\label{sec:intro}
Robotic manipulation of deformable objects is a challenging problem especially because of the
complexity of the many different ways an object can deform. Searching within such a high dimensional
state space makes it difficult to recognize, track, and manipulate deformable objects. In
this paper we present a feed-forward, model-driven approach to address this challenge, using a
pre-computed, simulated database of deformable thin-shell object models, where the bending of the mesh models is predominant  \cite{Grinspun:2003}. The models are detailed, robust,
and easy to construct, and using a physics engine one can  accurately  predict the behavior of the objects in simulation, which can then be applied to a real physical setting. This work bridges the gap between the
simulation world and the real world. The predictive, feed-forward, model-driven approach takes advantages
of the simulation and generates a large number of instances for learning approaches, which not
only alleviates the burden of data collection, which can be efficiently done in simulation,
but also makes adaptation of the methods to other application areas easier and faster.
Mesh models of common deformable garments are simulated with the garments
picked up in multiple different poses under gravity, and stored in a database for fast and
efficient retrieval. To validate this approach, we developed a comprehensive pipeline for manipulating
clothing as in a typical laundry task. First, the database is used to estimate categories and poses of garments
in arbitrary positions. A fully featured 3D volumetric model of the garment is constructed in
real-time and volumetric features are then used to obtain the most similar model in the database to
predict the object category and pose. Second, the database can significantly benefit the manipulation
of deformable objects via non-rigid registration, providing accurate correspondences between the
reconstructed object model and the database models. Third, the accurate model simulation can also
be used to optimize the trajectories for manipulation of deformable objects, such as the folding of
garments.  In addition, the simulation can be easily adapted to new garment models. Extensive experimental results are shown for the tasks above using a variety of different clothing.

\begin{figure}[t]
    \center
    \begin{tabular}{c}
        \includegraphics[width=0.48\textwidth]{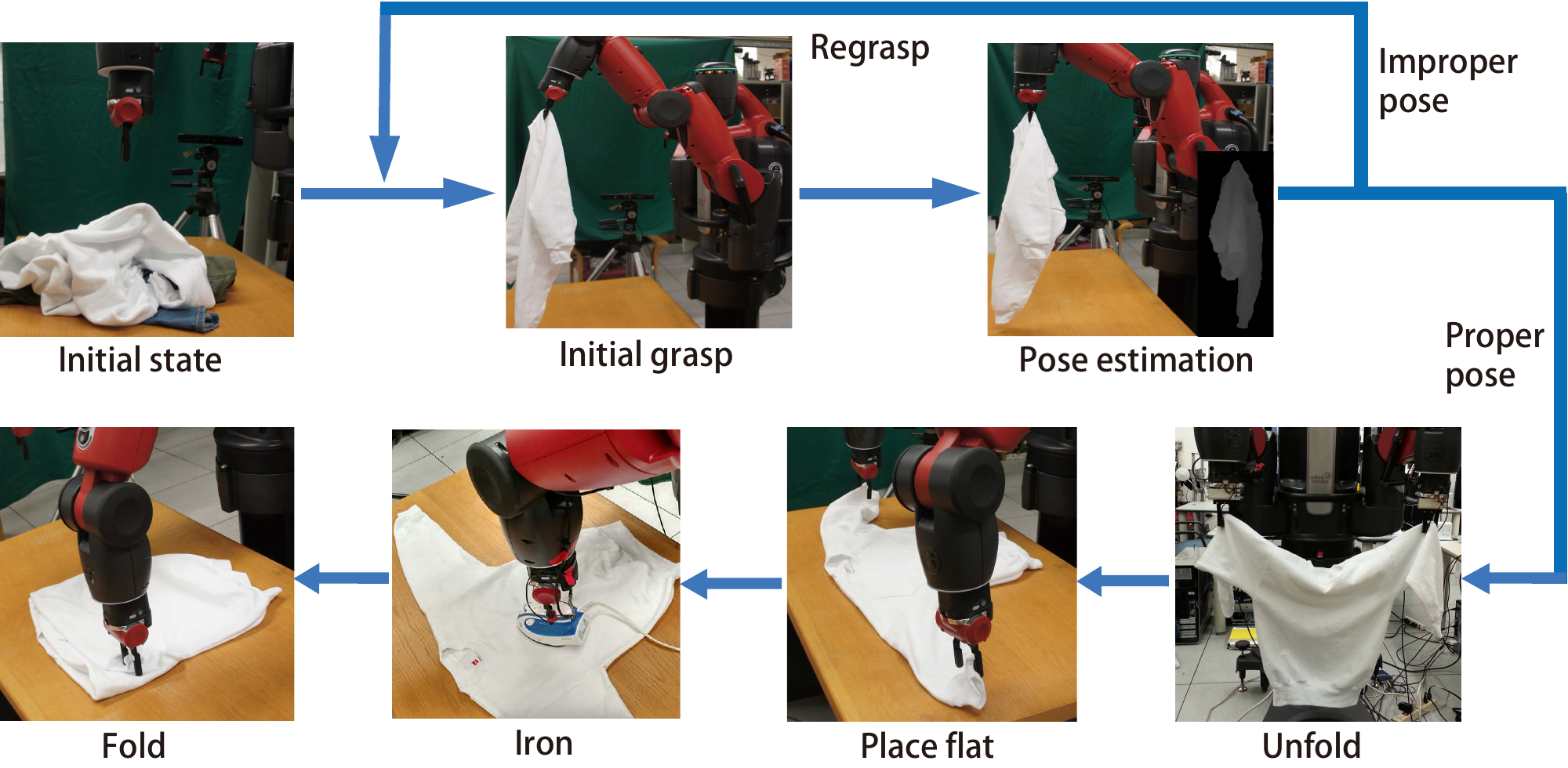}
    \end{tabular}
    \caption{The overall pipeline of robotic manipulation of deformable objects.}
    \label{fig:entire_pipeline}
\end{figure}

Figure \ref{fig:entire_pipeline} shows a typical pipeline for manipulating clothing as in a laundry task.  This paper brings together  work addressing all the tasks in the pipeline (except ironing) which have been previously published in conference papers (\cite{LiICRA2014}~\cite{LiIROS2014}~\cite{LiICRA2015}~\cite{LiIROS2015}). These tasks, with the exception of the ironing task, are all implemented using a feed forward, model-driven methodology, and this paper serves to consolidate all these results into a single integrated whole. The work has also been extended to include novel garments not found in the database, extended results on regrasping using a much larger dataset of objects and examples, quantitative registration results for our hybrid rigid/deformable registration methods, new dense mesh modeling techniques, and a novel dissimilarity metric used to assess folding success. 
The ironing task is omitted from this paper due to size constraints. Full details on ironing can be found in ~\cite{LiICRA2016}.

In addition, a set of videos of our experimental results are available at: \url{https://youtu.be/fRp05Teua4c}\section{Related Work}
\label{sec:relatedwork}

\subsection{Recognition}
There has been previous work on the recognition and manipulation of deformable objects. \cite{Willimon2013}~\cite{Willimon2011} used interactive perception to classify the clothing type. 
Their work was based on a image-only database of 6 categories, each of which is with 5 different items from real garments. 
Later, they increased the size of the database but still used real garments. 
Their work focused on small clothing such as socks and short pants usually consisting of a single color. 
\cite{Milleretal_IJRR2012}, \cite{Wangetal_IROS2011}, \cite{SchulmanLeeHoAbbeel_ICRA2013}, Cusumano-Towner, et. al~\cite{Towner2011} have done some impressive work in clothing recognition and manipulation. 
They have successfully enabled the PR2 robot to fold clothing and towels. Their methods mainly focus on aligning the current edge/shape from observation to an existing shape. 
A series of works on clothes pose recognition were done by ~\cite{Kita2011}~\cite{Kita2009}~\cite{Kita2002}. 
They used a simulation database of a single garment with around $18$ different grasping points, which were mostly selected on the border when the garment was laid flat.
Their work demonstrated the ability to identify the pose of the clothes by registration to pre-recorded template images. 
\cite{DoumanoglouICRA2014} used a pair of two industrial arms to recognize and manipulate deformable garments.
They used a database of depth images captured from $24$ real garments, such as a sweater or a pair of pants.

With powerful computing resources reconstructing a 3D model of the garment, and using that to search a pre-computed database of simulated garment models in different poses can be more accurate and efficient.
With the increasing popularity of Kinect sensor, there are various methods emerging in computer graphics such as KinectFusion and its variants~\cite{Newcombe2011}~\cite{Chen2013}~\cite{li2013}.
Although these methods have shown success in reconstructing static scenes, they do not fit our scenario directly where a robotic arm is rotating the target garment about a grasping point.
Therefore we first do a 3D segmentation to get the masks of the garment on the depth images, and then invoke KinectFusion to do the reconstruction.

Shape matching is another related and long-standing topic in robotics and computer vision.
On the 2D side, various local features have been developed for image matching and recognition~\cite{1993Huttenlocher}~\cite{2000Latecki}~\cite{lowe1999}, which have shown good performance on textured images.
Another direction is shape-context based recognition~\cite{belongie2002pami}~\cite{2010Toshev}~\cite{2004Tu}, which is better for handwriting and character matching.
On the 3D side, Wu \cite{2008Wu} and Wang \cite{2006Wang} have proposed methods to match patches based on 3D local features. 
They extract Viewpoint-Invariant Patches or the distribution of geometry primitives as features, based on which matching is performed. 
\cite{2001Osada}, \cite{2003Thayananthan}, and \cite{Frome04} apply 3D shape-context as a metric to compute similarities of 3D layout for recognition. 
However, most of the methods are designed for noise-free human-designed models, without the capability to match between the relatively noisy and incomplete mesh model produced by Kinect and the human-designed models.
Our method is inspired by 3D shape context~\cite{Frome04}, but provides the capability of cross-domain matching with a learned distance metric, and also utilizes a volumetric data representation to efficiently extract the features.

\subsection{Manipulation}

\cite{Osawa2007} proposed a method using a dual-arm setup to unfold a garment from pick up. 
They used a segmented mask to match the pre-stored template mask to track the states of the garment.
The PR2 robot is probably the first robot that has successfully manipulated deformable objects such as a towel or a T-shirt~\cite{Shepard2010}. 
The visual recognition in this work targets corner-based features, which does not require a template to match.
The subsequent work has improved the prediction of the state of a garment using a HMM framework by regrasping at the lowest corner point~\cite{Towner2011}. 
\cite{DoumanoglouICRA2014} applied pre-recorded depth images to guided the manipulation procedures. \cite{sun2015} used a pair of stereo cameras to analysis the surface of a piece of cloth and performed flattening and unfolding.

One of the applications of the our database is to localize the regrasping point during the manipulation by mapping the pre-determined points from simulation mesh to the reconstructed mesh.
Therefore, a fast and accurate registration algorithm plays a key role in our method.
Rigid or non-rigid surface registration is a fundamental tool to find shape correspondence.
A thorough review can be found in~\cite{Tam:2013}.
Our registration algorithm builds on previous techniques for rigid and non-rigid registrations.
First, we use an iterative closest point method~\cite{Besl:1992} to rigidly align the garment.
Here, we use a distance field to accelerate the computation.
Next, we perform a non-rigid registration to improve the matching by locally deforming the garment.
Similar to~\cite{Li:2008}, we find the correspondence by minimizing an energy function
that describes the deformation and the fitting.

\subsection{Folding Deformable Objects}

With the garment fully spread on the table, attention is turned to parsing its shape.
S. Miller~\etal have designed a parametrized shape model for unknown garments~\cite{millerICRA2011}~\cite{Milleretal_IJRR2012}.
Each set of parameters defines a certain type of garment such as a sweater or a towel.
The goal is to minimize the distance between the observed garment contour points and points from the parametrized shape model.
The fitting score between the observed contour and the shape models can also be used for recognition of garment category.
However, the average time for the fitting procedure is $30-150$ seconds and sometimes does not converge.
The contour-based garment shape model was further improved by J. Stria~\etal using polygonal models~\cite{Stria2014TAROS}.
The detected garment contour is matched to a polygonal model by removing non-convex points using a dynamic programming approach.
The landmarks on the polygonal model are then mapped to the real garment contour, and followed by generating a folding plan.

Folding is another application of garment manipulation.
F. Osawa \etal used a robot to fold a garment with a special purpose table that contains a plate that can bend and fold the clothes assisted by a dual-arm robot.
The robot mainly worked on repositioning the clothes for the plate for each folding action.
Within several ``flip-fold'' operations, the garment can be folded.
Another folding method using a PR2 robot was implemented by \cite{vandenberg2010}.
The core of their approach was the geometry reasoning with respect to the cloth model without any physical simulation.
Contour fitting at each step took relatively longer than execution of the folding actions, which reduced its efficiency.
This was further sped up by \cite{StriaIROS2014} using two industrial arms and a polygonal contour model.
They showed impressive folding results by utilizing a specifically-designed gripper~\cite{Le2013} that is suitable for cloth grasping and manipulation.

None of the previous works focus on trajectory optimization for garment folding, which brings uncertainty to the layout given the same folding plan.
One possible case is that the garment shifts on the table during one folding action so that the targeted folding position is also moved.
Another case is that an improper folding trajectory causes additional deformation of the garment itself, which can accumulate.
Our previous work~\cite{LiIROS2015} has proved that with effective simulation, bad trajectories can be avoided and the results of manipulation of the deformable objects is predictable.

\section{A Database For Deformable Object Recognition}
\label{sec:database}

\subsection{Motivation}
Figure~\ref{fig:entire_pipeline} shows an overview of our pipeline for dexterous manipulation of deformable objects.
The first step is the visual recognition of deformable objects. 
We need to have a large set of exemplars of how garments will look visually when arbitrarily grasped. 
In addition, as we mentioned previously, a 3D model can be used for regrasping and further manipulation after an accurate registration.
Therefore, a database with a number of 3D models is desirable.
In order to have a set of pre-calculated trajectories for efficient manipulation of deformable objects, off-line simulation is an effective way to approach this.
With low-cost and fast simulation, optimized trajectories can be calculated, and exported and adapted to the real robotic manipulation.

Physically having people, or a robot arm, successively pick up an object and image its appearance is too slow and cannot span the large space we are hoping to learn. 
Given the physical nature of this training set, it can be very time-consuming to create, and may have problems encompassing a wide range of garments and different fabrics which we can more easily accommodate in the simulation environment.
Using advanced simulators such as \emph{Maya}~\cite{Maya2015} to physically simulate deformable objects, we can produce thousands of exemplars efficiently, which we can then use as a corpus for learning the visual appearances of the deformed garments.

Take manipulation of the deformable garment as an example.
One solution is to use prior knowledge to guide the robot to follow steps of a task.
In our previous work, we have successfully used online registration between the database model and the reconstructed model to achieve a stable regrasping and unfold a garment.
The work that is closest to ours is by ~\cite{DoumanoglouICRA2014}.
This work has impressive results for unfolding of a number of different garments.
They use a dual-arm industrial robot to unfold a garment guided by a set of depth images which provide a regrasping point. This method achieves promising accuracy.
Their training set is a number of physical garments that have been grasped at different grasping points to create feature vectors for learning.
A major difference is our use of simulated predictive thin shell models for garments from a large database of garments and their poses.
We also use an online registration of a reconstructed 3D surface mesh with the simulated model to find regrasping points.
By this method, we can choose arbitrary regrasping points without having to train the physical model for the occurrence of the grasping points.
This allows us to choose any point on the garment at any time as the regrasping point.

\subsection{Simulating Deformable Objects}
\begin{figure}[!ht]
\center
\begin{tabular}{c}
\includegraphics[width=0.45\textwidth]{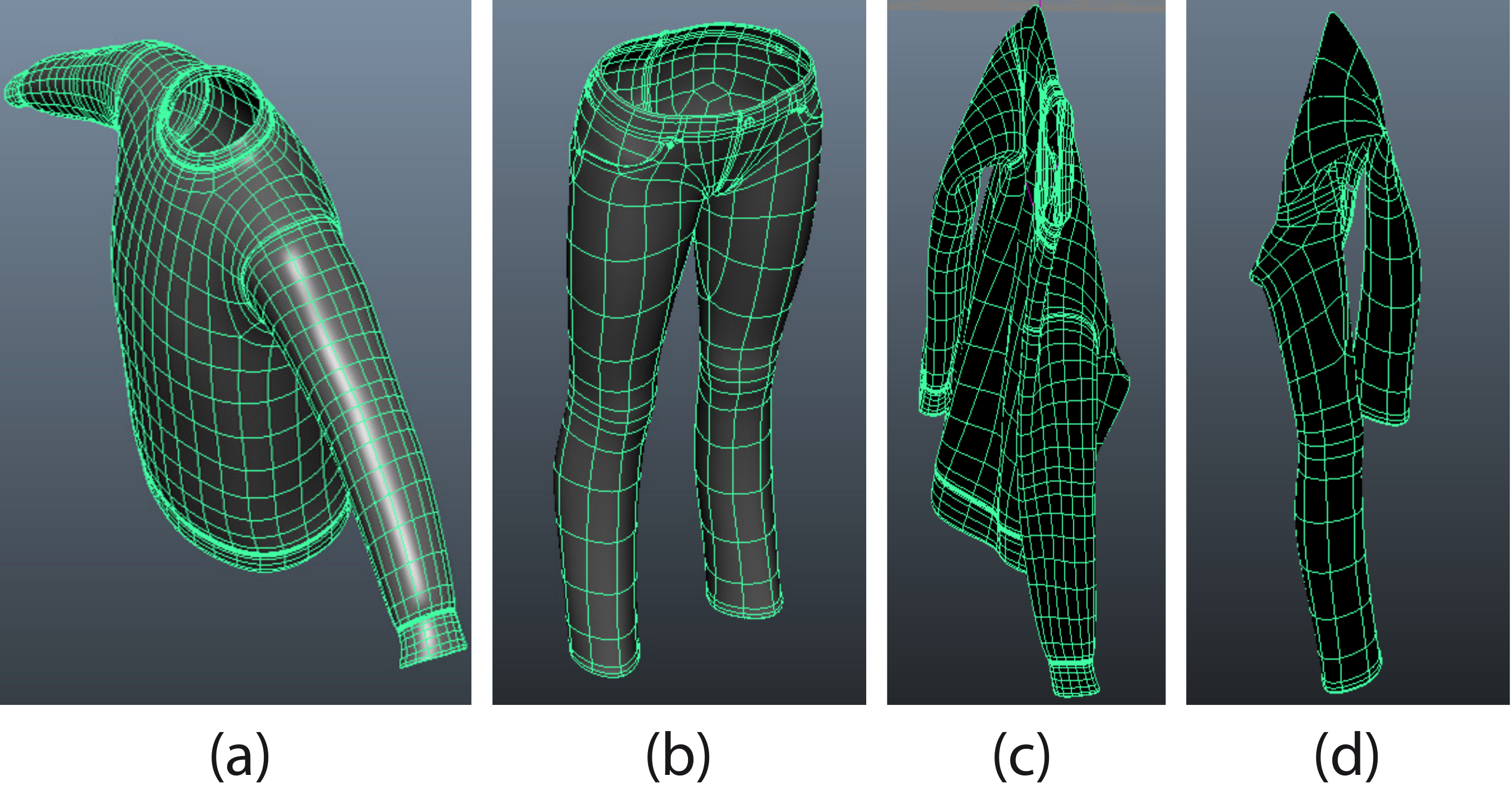}
\end{tabular}
\caption{{(a), (b): } The original garment mesh models of a sweater and a pair of jeans rendered in \emph{Maya}. {(c), (d): } Simulation result of hanging (a) and (b) under under gravity, respectively. }
\label{fig:ModelFromMaya}
\end{figure}

We have developed an off-line simulation pipeline whose results can be used to predict poses of deformable objects. 
The off-line simulation is time efficient, noise free, and more accurate compared with acquiring data via sensors from real objects. Simulation models do not suffer from occlusion or noise and are more complete than physically scanned models.
In the off-line simulation, we use a few well-defined garment mesh models such as sweaters, jeans, and short pants, etc. 
Similar garment mesh models can also be obtained from Poserworld~\cite{PoserWorld} and Turbo Squid~\cite{TurboSquid}.  
We can also generate models by using our own \lq\lq \emph{Sensitive Couture} \rq\rq~software~\cite{CG_0179}.  
Figure~\ref{fig:ModelFromMaya} shows a few of our current garment models rendered in \emph{Maya} software.

For each grasping point, we compute the garment layout by hanging under gravity in the simulator.
In \emph{Maya}, a mesh model can be converted into an \emph{nCloth} model which can be then simulated with some cloth properties such as hanging and falling down. 
\emph{Maya} also allows for control of cloth thickness and deformation resistance, etc.
In addition, any vertex on the mesh can be selected as a constraint point to simulate a draping effect. 
The hanging under gravity effect of the garment models is shown in Figure~\ref{fig:ModelFromMaya}.
We manually label each garment in the database with the key grasping points such as sleeve end, elbow, shoulder, chest, and waist, etc.

\begin{figure}[htb]
\center
\begin{tabular}{c}
 \includegraphics[width=0.48\textwidth]{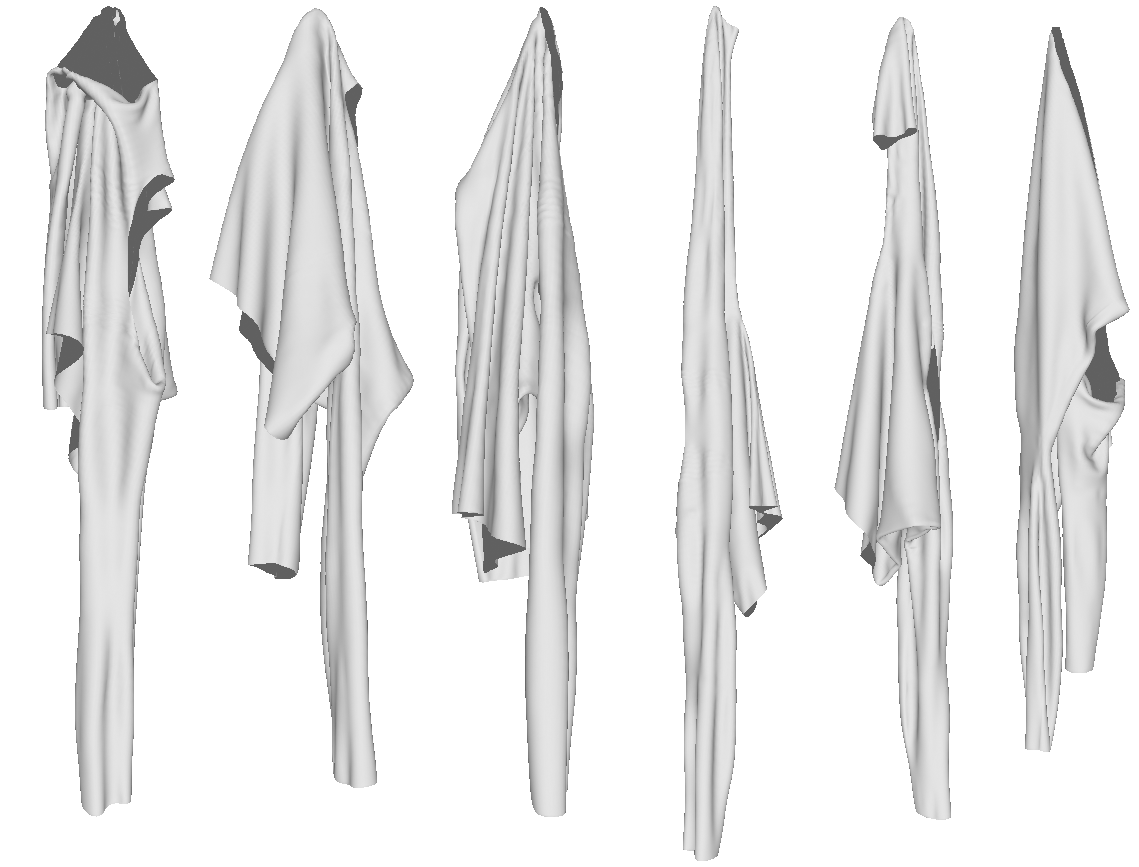}
\end{tabular}
\caption{Six different mesh models of a same sweater, but picked up on different points, simulated in \emph{Maya}.}
\label{fig:sample_picking_simulation}
\end{figure}

The simulation model be can exported as an obj file for recognition using volumetric approach~\cite{LiIROS2014}.
Figure~\ref{fig:sample_picking_simulation} shows a small sample of different picking points of a single garment hanging under gravity that simulated in \emph{Maya}.

\section{Pose Estimation}
\label{ch:recognition_volumetric}

Pose estimation of deformable objects is an important problem in robotics, laying the foundation for further procedures.  For example, in the task of garment folding, once the robot has detected the pose of the garment, it can then proceed to manipulate the target garment to a preset ``standard pose.''
Unlike rigid object recognition which has finite state spaces, deformable object recognition is much harder because of the very large state space of how it deforms.

\begin{figure}[!htbp]
    \center
    \includegraphics[width=0.48\textwidth]{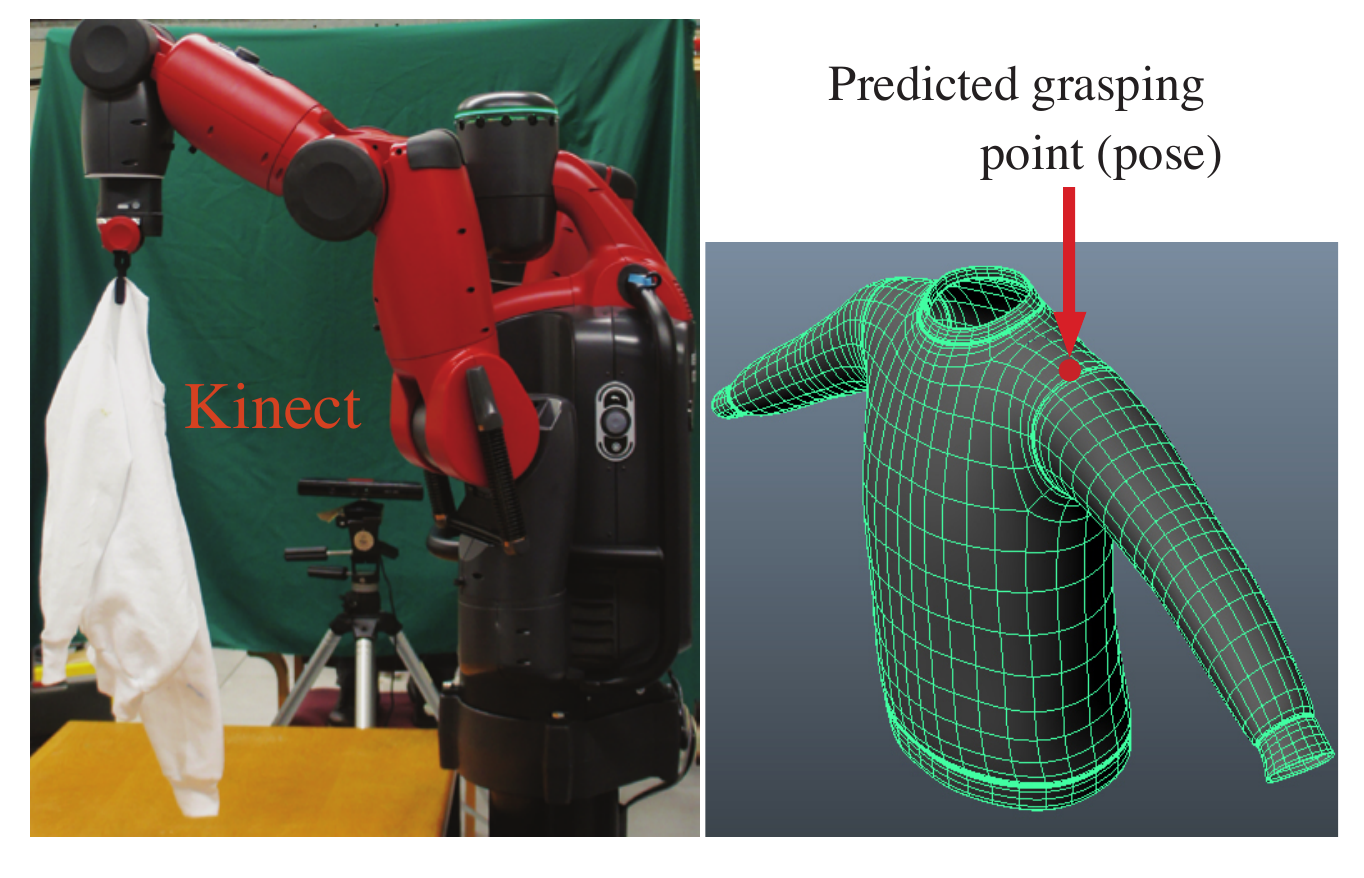} 
    \caption{
        Our application scenario: a Baxter robot grasps a sweater, and a Kinect captures depth images to recognize the pose of the sweater.
        The recognition result is shown on the right.
    }
    \label{fig:scenario_pose}
\end{figure}
In this section, we describe a real-time pose recognition algorithm with accurate prediction of grasping point locations. 
Figure~\ref{fig:scenario_pose} shows the experimental settings for our algorithm: a Baxter robot grasping a garment and predicting the grasping location (e.g. $2$cm left of the collar).
With this information, the robot is then able to proceed to subsequent tasks such as regrasping and folding.
The main idea of our method is to first accurately reconstruct a 3D mesh model from a low-cost depth sensor,
and then compute the similarity between the reconstructed model and the models simulated offline to predict the pose of the object. 
The database introduced in the previous section provides a perfect source for such offline-simulated models.

\begin{figure*}[!htpb]
    \centering
    \includegraphics[width=1.0\textwidth]{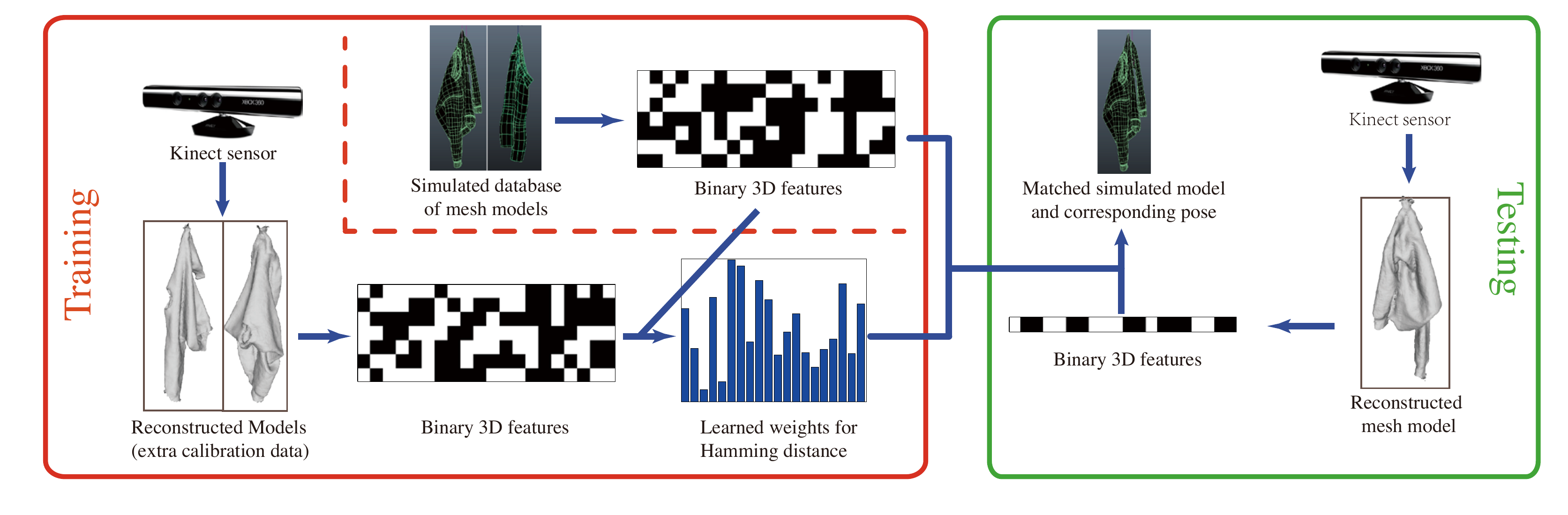}
    \caption{ 
        Overview of our proposed pipeline for pose estimation of deformable objects.
        In the offline training stage (the red rectangle), we extract a tailored binary feature from the simulated database, and learn a weighted Hamming distance from additional calibrated data collected from the Kinect.
        In the online testing stage (the green rectangle), we reconstruct a 3D model from the depth input, find the nearest neighbor from the simulated database with the learned distance metric, and then adopt the pose of the matched model as the output. 
    }
    \label{fig:pose_flowchart}
\end{figure*}

\subsection{Method}
\label{section:poseestimation_approach}

Our method consists of two stages: the offline model simulation stage and the online recognition stage.
In the offline model simulation stage, we use a physics engine to simulate the stationary state of the mesh models of different types of garments in different poses.
In the online recognition stage, we use a Kinect sensor to capture many depth images of different view points of the garment by rotating it as it hangs from a robotic arm.
We then reconstruct a smooth 3D model from the depth input, extract compact 3D features from it, and finally match against the offline model database to recognize its pose.
Figure~\ref{fig:pose_flowchart} shows the framework of our method, which will be introduced in the subsequent subsections.

\subsubsection{3D Reconstruction}
\label{subsec:reconstruction}

Given the model database described above, we now need to generate depth images and match against the database. 
Direct recognition from depth images suffers from the problems of self-occlusion and sensor noise.
This naturally leads to our new method of first building a smooth 3D model from the noisy input, and then performing recognition in 3D.
However, how to do such reconstruction is still an open problem. Although there are existing approaches of obtaining high-quality models from noisy depth inputs such as KinectFusion~\cite{Newcombe2011}, which requires the scene to be static.
In our data collection settings, the target garment is being rotated by a robotic arm, which invalidates the KinectFusion's assumptions.
We solve this problem by first segmenting out the garment from its background, and then invoke KinectFusion to obtain a smooth 3D model, assuming that the rotation is slow and steady enough such that the garment will not deform in the process.

\textbf{Segmentation.}
Before diving into the reconstruction algorithm, let us first define some notation.
Given the intrinsic matrix $F_d$ of the depth camera and the $i$th depth image $I_i$, we are able to compute the 3D coordinates of all the pixels in the camera coordinate system with $\begin{bmatrix} x_{ci} & y_{ci} & z_{ci} \end{bmatrix}^T = F^{-1}d_i \begin{bmatrix}u_i & v_i & 1 \end{bmatrix}^T$, 
in which $(u_i, v_i)$ is the coordinate of a pixel in $I_i$, with $d_i$ as the corresponding depth, and $(x_{ci},y_{ci},z_{ci})$ is the corresponding 3D coordinate in the camera coordinate system.

Our segmentation is then performed in the 3D space.
We ask the user to specify a 2D bounding box on the depth image $(x_{\min}, x_{\max}, y_{\min}, y_{\max})$ with a rough estimation of the depth of the garment $(z_{\min}, z_{\max})$.
Given that the data collection environment is reasonably constrained, we find even one predefined bounding box works well.
Then we adopt all the pixels having their 3D coordinates within the bounding box as the foreground, resulting in a series of masked depth images $\{I_i\}$ and their corresponding 3D points, which will be fed into the reconstruction module.

The 3D reconstruction is done by feeding the masked depth images $\{I_i\}$ into KinectFusion, while the unrelated surroundings are eliminated, leaving the scene to reconstruct as static.
This process can be done in real time.
In addition to a smooth mesh, the KinectFusion library also generates a Signed Distance Function (SDF) mapping, which will be used for 3D feature extraction. 
The SDF is defined on any 3D point $(x,y,z)$.
It has the property that it is negative when the point is within the surface of the scanned object, positive when the point is outside a surface, and zero when it is on the surface.
We will use this function to efficiently compute our 3D features in the next subsection.

\subsubsection{Feature Extraction}
\label{subsec:feature_extraction}

\begin{figure}[!htpb]
    \center
    \includegraphics[width=0.50\textwidth]{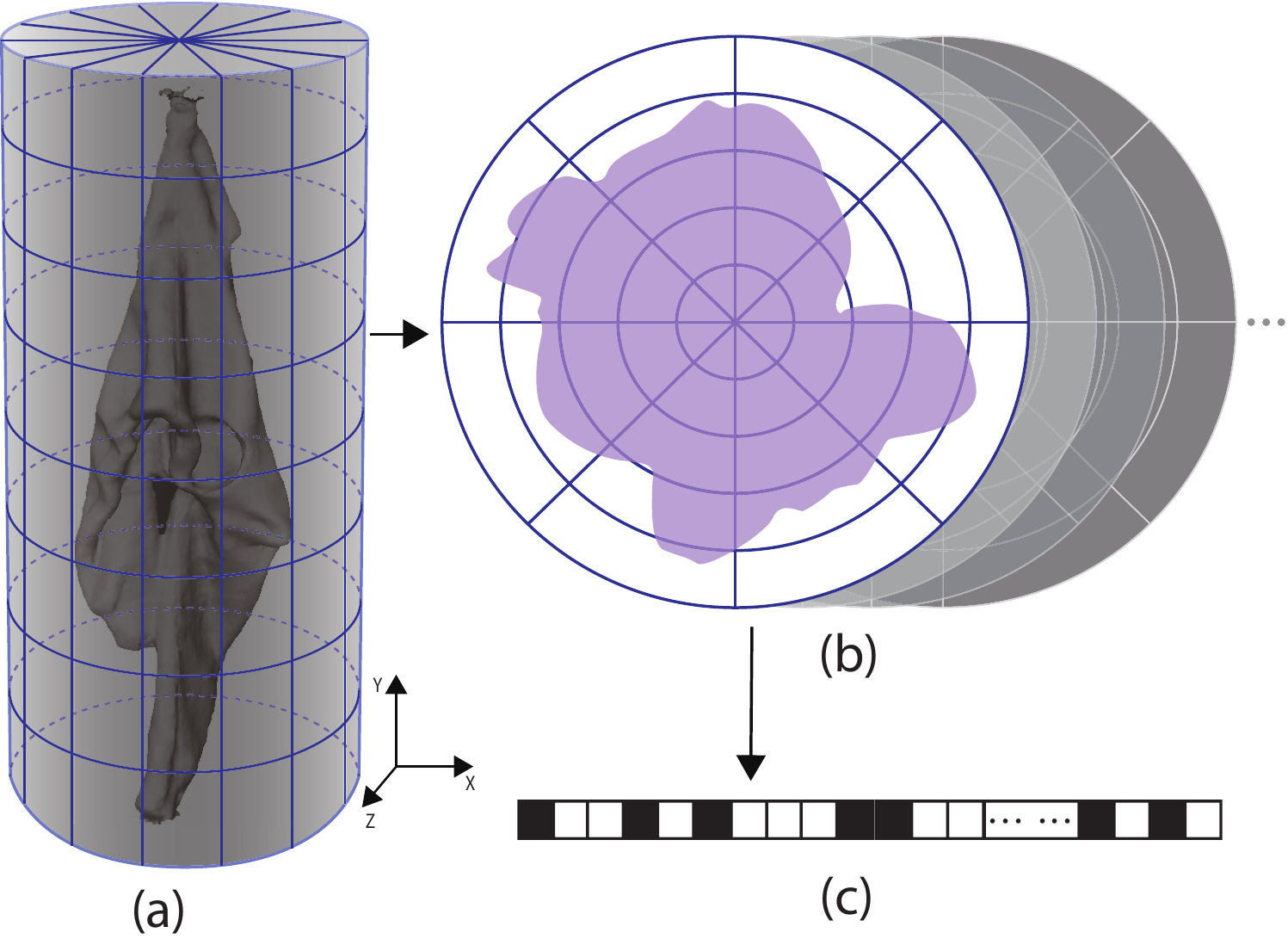}
    \caption{Feature extraction from a reconstructed mesh model. 
        (a) indicates that a bounding cylinder of a garment is cut into several layers.
        (b) shows a set of layers (sections). For each layer, we divide it into cells via rings and sectors. 
        (c) shows a binary feature vector collected from each cell. 
        Details are described in section~\ref{subsec:feature_extraction}. }
        \label{fig:feature_extraction}
    \end{figure}

    Inspired by 3D Shape Context~\cite{belongie2002pami}, we design a binary feature to describe the 3D models. 
    In our method, the features are defined on a cylindrical coordinate system fit to the hanging garment as opposed to traditional 3D Shape Context which uses a spherical coordinate system~\cite{Frome04}.

    For each layer, as shown in Figure~\ref{fig:feature_extraction} top-right, we \emph{uniformly} divide the world space into $(R$ rings$) \times (\Phi$ sectors$)$ in a polar coordinate system, with the largest ring covering the largest radius among all the layers.
    The center of the polar coordinate system is determined as the mean of all the points in the highest layer, which usually contains the robot gripper.
    Note we do a uniform division instead of logarithm division of $r$ as Shape Context does. 
    The reason why Shape Context uses logarithm division of $r$ is that the cells farther from the center are less important, which is not the case in our settings.
    For each layer, instead of doing a point count as in the original Shape Context method, we check the Signed Distance Function (SDF) of the voxel which the center of the polar cell belongs to, and fill one ($1$) in the cell if the SDF is zero or negative (i.e. the cell is inside the voxel), otherwise zero ($0$).
    Finally, all the binary numbers in each cell are collected in an order (e.g. with $\phi$ increasing and then $r$ increasing), and are concatenated as the final feature vector.

    The insight behind this design is, 
    to improve the robustness against local surface disturbance due to friction, we include the 3D voxels \emph{inside} the surface in the features.
    Note we do not need to do the time-consuming classification (e.g. ray tracing) to determine whether each cell is inside the surface, but only need to look up their SDFs, thus dramatically speed up the feature extraction.

    \begin{algorithm}[!htpb]
        \caption{Feature extraction for pose estimation of deformable objects}
        \label{alg:feat}
        \KwIn{                      
            Vertices of the mesh model $\Omega = \{v_i\}$, 
            precomputed SDF$(x, y, z)$,
        Parameters: $N$ = \#layers, $R$ = \#rings, $\Phi$ = \#sectors}
        \KwOut{Corresponding feature vector ${{\bm x}} \in \mathbb{B}^{lrs}$}
        ${\bm x} = {\bm 0} \in \mathbb{B}^{lrs}$\;
        Divide mesh $\Omega$ into $l$ layers $\Omega_1, \Omega_2 \cdots $ in a top-down manner\;
        Origin = Mean$(\Omega_{1x}, \Omega_{1z})$\;
        [${\bm r}, {\bm \phi}$] = Polar (Origin, $\Omega_{x}, \Omega_{z}$) \;
        $r_m = \max {\bm r}$\;

        \For{each layer $\Omega_i$} {
            \For{each cell (ring, sector) $\in \Omega_i$} {
                $(x, y, z) = $ center of the cell \;
                \If{SDF$(x, y, z) \leq 0$} {
                    ${\bm x}\left[\frac{rR\Phi}{r_m} + \frac{\phi\Phi}{2\pi} \right] = 1$ \;
                }\Else{
                    ${\bm x}\left[\frac{rR\Phi}{r_m} + \frac{\phi\Phi}{2\pi} \right] = 0$ \;
                }
            }
        }	
        \Return{${\bm x}$}.
    \end{algorithm}

    \textbf{Matching Scheme.}
    Similar to Shape Context, when matching against two shapes, we conceptually rotate one of them and adopt the minimum distance as the matching cost, to provide rotation invariance.
    That is,
    \begin{equation}
        \text{Distance}({\bm x}_1,{\bm x}_2) = \min_i \lVert R_i {\bm x}_1 \oplus {\bm x}_2 \rVert_1,
        \label{eq:match}
    \end{equation}
    in which ${\bm x}_1, {\bm x}_2 \in \mathbb{B}^{\Phi R N}$ are the features to be matched ($\mathbb{B}$ is the binary set $\{0, 1\}$), $\oplus$ is the binary XOR operation, and $R_i$ is the transform matrix to rotate the feature of each layer by $2\pi/\Phi$.
    Recall that both features to be matched are compact binary codes. Thus such conceptual rotation as well as Hamming distance computation can be efficiently implemented by integer shifting and XOR operations, resulting in matching that is even faster than the Euclidean Distance given reasonable $\Phi$s (e.g. $\Phi = 10$). A complete illustration of the feature extraction algorithm can be found in \Alg{feat}.

    \subsubsection{Domain Adaptation}
    \label{subsec:distance}

    Now we have a feature vector representation for each model in the simulated database and for the query.
    A natural idea is to find the Nearest Neighbor (NN) of the query in the database and transfer the metadata such as category and pose from the NN to the query.
    But a naive NN algorithm with Euclidean distance does not work here because， 
    even for the same garment and the same grasping point by the robot, the way it deforms may still be slightly different due to friction.
    This requires a solution in the matching stage, especially given that it is impractical to simulate every object with all the possible materials.
    Therefore, essentially we are doing cross-domain retrieval, which generally requires a ``calibration'' step to adapt the knowledge from one domain (simulated models) to another (reconstructed models).

    \textbf{Weighted Hamming Distance.}
    Similar with the distance calibration in \cite{Wang2013}, we use a \emph{learned} distance metric to improve the NN accuracy, i.e.
    \begin{equation}
        \text{BestMatch}_{\bm w}({\bm q}) = \arg \min_{i} {\bm w}^T \left( \hat{{\bm x}}_i \oplus {\bm q} \right),
        \label{eq:bestmatch}
    \end{equation}
    in which ${\bm q}$ is the feature vector of the query, $i$ is the index of models in the simulated database, and $\oplus$ is the binary XOR operation.
    ${\bm \hat{x}}_i = \hat{R}_i{\bm x}_i$ indicates the feature vector of the $i$th model, with $\hat{R}_i$ as the optimal $R$ in \Eq{match}.

    The insight here is that we wish to grant our distance metric more robustness against material properties by assigning larger weights to the regions invariant to the material differences (this amplifies the features that are more intrinsic for the recognition task).

    \textbf{Distance Metric Learning.}
    To robustly learn the weighted Hamming distance, we use an extra set of mesh models collected from a Kinect as \emph{calibration data}.
    The collection settings are the same as described in ``3D Reconstruction'' and only a small amount of calibration data is needed for each category (e.g. $5$ models in $19$ poses for long-sleeve shirt model).
    To determine the weight vector ${\bm w}$, we then formulate the learning process as an optimization problem of minimizing the empirical error with a large-margin regularizer:
    \begin{equation}
        \begin{aligned}[rl]
            \min_{\bm w} & \frac{1}{2} \lVert {\bm w} \rVert_2^2 + C \sum_j \xi_j \\
            \text{s.t. } & {\bm w}^T \left( \hat{{\bm x}}_i \oplus {\bm q}_j \right) < {\bm w}^T \left( \hat{{\bm x}}_k \oplus {\bm q}_j \right) + \xi_j , \\
            & \forall j, \forall y_i = l_j, y_k \neq l_j, \\
            & \xi_j \geq 0,
        \end{aligned}
        \label{eq:learn}
    \end{equation}
    in which ${\bm \hat{x}}_i$ is the orientation-calibrated feature of the $i$th model (from the database), with $y_i$ as the corresponding ground truth label (i.e. the index of the pose).
    ${\bm q}_j$ is the extracted feature of the $j$th training model (from Kinect), with $l_i$ as the ground truth label.
    We wish to minimize $\sum_i \xi_i$, which indicates how many wrong results the learned metric ${\bm w}$ gives, with a quadratic regularizer.
    $C$ controls how much penalty is given to wrong predictions.

    This is a non-convex and even non-differentiable problem.
    Therefore we employ the RankSVM~\cite{Joachims2002} to obtain an approximate solution using the cutting-plane method.

    \textbf{Knowledge Transfer.}
    Given the learned ${\bm w}$, in the testing stage, we then use \Eq{bestmatch} to obtain the nearest neighbor of the query model.
    We directly adopt the grasping point of the nearest neighbor, which is known from the simulation process, as the final prediction.

    \subsection{Experimental Results}
    \label{section:poseestimation_experiments}

    We used a series of experiments to demonstrate the effectiveness of the proposed method and justify the components.
    We tested our method on a dataset of various kinds of garments collected from practical settings, by treating it as a classification problem and calculating the classification accuracy.
    Experimental results demonstrate that our method is able to achieve both reasonable accuracy and fast speed.

    \subsubsection{Data Acquisition}

    Since the simulated database introduced in the previous section does not have the practically captured data, we collect an extra test dataset for general evaluation of pose recognition of deformable objects based on depth image as inputs. 
    \begin{figure*}[!htpb]
        \begin{tabular}{c}
            \hspace{-0.4cm}
            \includegraphics[width=1\textwidth]{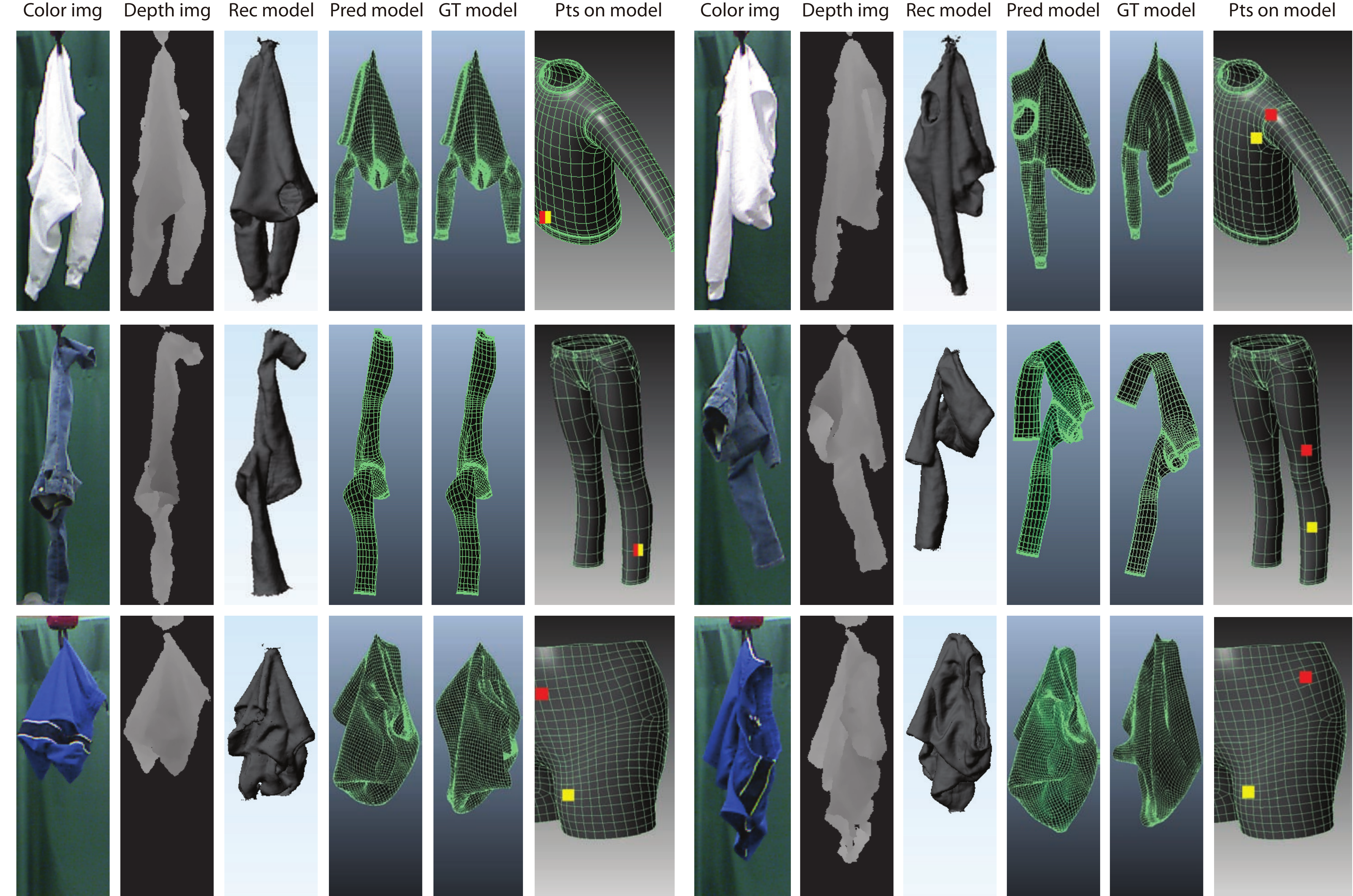}
        \end{tabular}
        \caption{
            Visual examples of the pose recognition result of our method.
            The garment is picked up via a griper of the Baxter robot.
            From left to right, each example shows the color image, input depth image, reconstructed model, matched simulated model, ground truth simulated model, and the predicted grasping points (red) marked on the model with the ground truth (yellow).
            The example shown in the bottom right shown here is considered as a failure example, which may be because of the uninformative deformation shape.
            Note our method does not use any color information.
            (Best viewed in color)
        }
        \label{fig:expExample}
    \end{figure*}

    The dataset consists of 2 parts, a test set and a calibration set.
    To collect the testing set, we use a Baxter robot, which is equipped with two arms with 7 degrees of freedom. 
    A Kinect sensor is mounted on a horizontal platform at height of $1.2$ meters to capture the depth images, as shown in \Fig{scenario_pose}.
    We bought $3$ kinds of garments -- long-sleeve shirts, pants and shorts, as representative examples in the manufacturing industry, and then collect their depth images with the same grasping points of the training database.
    We then use our 3D reconstruction algorithm to obtain their mesh models.
    For each grasping point of each garment, the robot rotates the garment 360 degrees around $10$ seconds while the Kinect captures at $30$fps, which gives us around $300$ depth images for each garment/pose.
    This results in a test set of $39$ mesh models, with their raw depth images.

    Given we also need to learn/calibrate a distance metric from extra data from Kinect (using \Eq{learn}), we collect an extra small amount of data with the same settings as the calibration data, only collecting five poses for each garment. 
    A weight vector ${\bm w}$ is then learned from this calibration data for each type of garment.

    \subsubsection{Qualitative Evaluation}

    We demonstrate some of the recognition results in \Fig{expExample} in the order of color image, depth image, reconstructed model, predicted model, ground truth model, and predicted grasping point (red) vs. ground truth grasping point (yellow) on the garment.
    From the figure, we can first see that our 3D reconstruction is able to provide us with good-quality models for a fixed camera capturing a dynamic scene.
    And our shape retrieval scheme with learned distance metrics is also able to provide reasonable matches for the grasping points.
    Note that our method is able to output a mesh model of the target garment, which is critical for the subsequent operations such as path planning and object manipulation.

    \subsubsection{Quantitative Evaluation}
    
     \mbox{}

    \textbf{Implementation Details.}
    In the 3D reconstruction, we set $X=384$, $Y=Z=768$ voxels and the resolution of the voxels as $384$ voxels per meter to obtain a trade-off between resolution and robustness against sensor noise.
    In the feature extraction, our implementation adopts $R=16, \Phi=16, N=16$ in the feature extraction as an empirically good configuration.
    That is, each mesh model gives a $16\times16\times 16=4096$ dimensional binary feature.
    We set the penalty $C=10$ in \Eq{learn}.

    \textbf{Classification Accuracy}.
    For each input garment, we compute the classification accuracy of pose recognition, i.e.

    \begin{equation}
        \text{Accuracy} = \frac{\text{\# of correctly classified test cases}}{\text{\# of all test cases}}
        \label{eq:accuracy}
    \end{equation}.

    The classification accuracy for each garment type is reported in \Tbl{accuracy} (left).
    Given we have two models for each garment in the database (except shorts), we report the accuracy achieved of using only Model 1 for retrieval, using only Model 2 for retrieval, and use all the available data.
    The total grasping points for long-sleeve shirts, pants, and shorts are $19$, $12$, and $8$ respectively. 
    Our method is benefited from the 3D reconstruction step, which reduces the sensor noise and integrates the information of each frame to a comprehensive model and thus leads to better decisions. 
    Among three types of garments, recognition of shorts is not as accurate as the other two. 
    One possible reason is that many of the shapes from different grasping points look very similar. 
    Even for human observers, it is hard to distinguish them. 

    \begin{table*}[t] 
        \begin{minipage}[]{0.55\textwidth} 
            \centering 
            \begin{tabular}{c|c|c|c} \hline
                Garment & Model 1 & Model 2 & Both models \\ \hline
                Long-Sleeve Shirts & $63.1\%$ & $68.4\%$ & $73.7\%$ \\
                Pants & $75.0\%$ & $75.0\%$ & $75.0\%$  \\
                Shorts & $62.5\%$ & N/A & $62.5\%$ \\
                \hline
            \end{tabular}
        \end{minipage}
        \begin{minipage}[]{0.45\textwidth} 
            \centering
            \begin{tabular}{c|c} \hline
                Garment & Running Time \\ \hline
                Long-Sleeve Shirts & $0.30$ \\
                Pants & $0.20$ \\
                Shorts & $0.22$ \\
                \hline
            \end{tabular}
        \end{minipage} 
        \caption{{\sc{Left:} }
        Average classification accuracy for different garment types. 
        {\sc{Right:}} 
        Average running time in seconds to process one garment of the proposed method on the proposed database, with the input of different garment types.
    }
    \label{tbl:accuracy}
\end{table*}

\textbf{Running Time.}
In addition, we also report the processing time of our method.
The time is measured on a PC with an Intel i7 3.0 GHz CPU, and shown in Table~\ref{tbl:accuracy} (right). 
We can see that our method demonstrates orders of magnitude speed-up against the state-of-the-art depth-image based method which takes minutes to process one input.
This verifies our advantages from the efficient 3D reconstruction, feature extraction, and matching.

\subsubsection{Generality to Novel Garments}

\begin{figure}[!htpb]
    \centering
    \begin{tabular}{c}
        \hspace{-0.5cm}
        \includegraphics[width=0.50\textwidth]{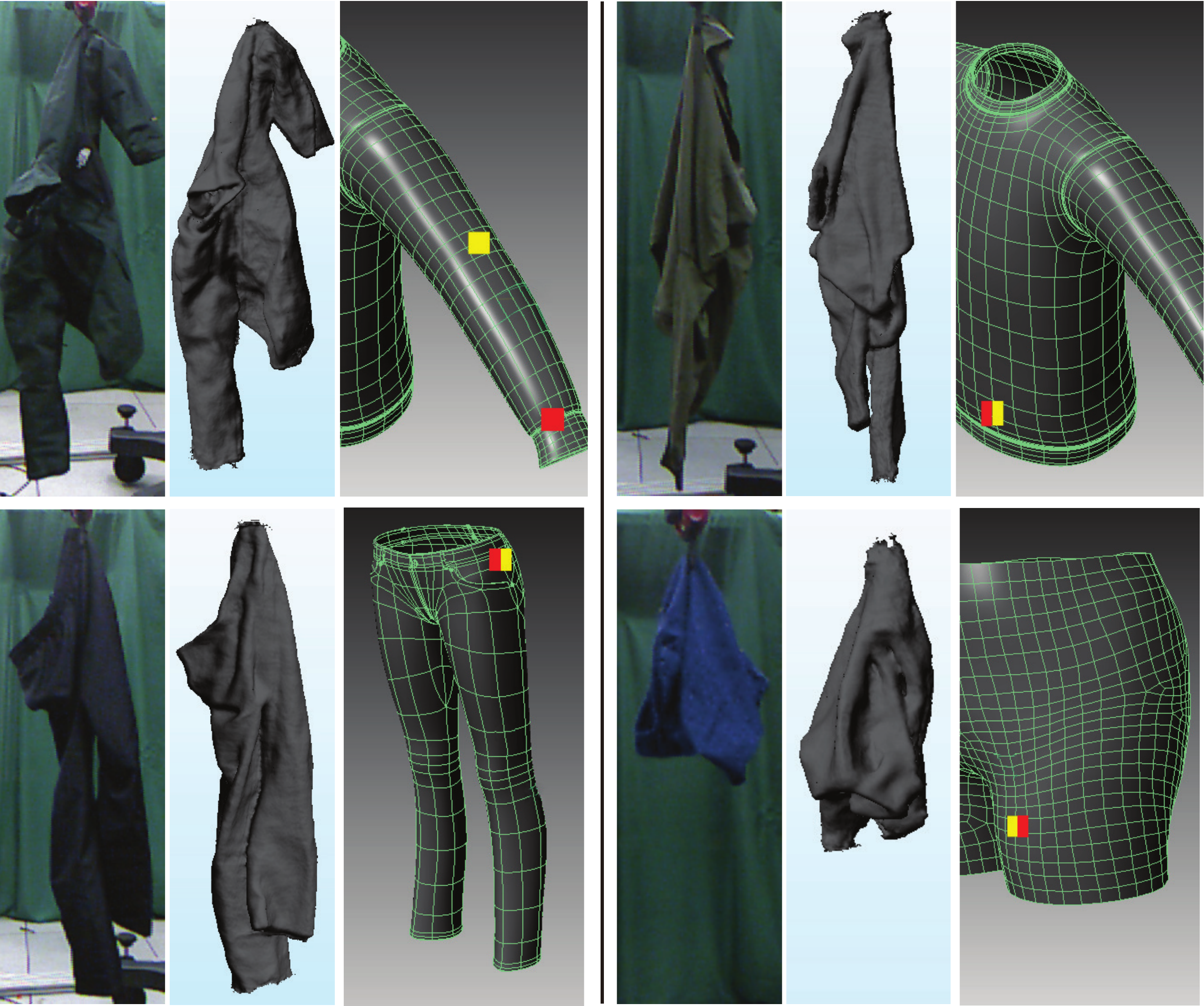}
    \end{tabular}
    \caption{
        Sample results of applying our method on novel garments.
        Each group of results shows the color image, reconstructed model, predicted grasping points (red) vs. ground truth (yellow) marked on the model from left to right.
        (Best viewed in color)
    }
    \label{fig:extra_Example}
\end{figure}

Though we used a relatively small garment database for our experiments, we noticed that our simulated models can also be generalized to recognize similar but unseen garments. 
For example, long-sleeve shirts and jackets can be considered as similar garments to our long-sleeve shirts model. 
Also, knit pants and suit pants are similar to our jeans model. 
Although they are made of different materials, the way they deform are similar to our training models in some poses.
Figure~\ref{fig:extra_Example} shows some extra examples of recognizing poses of unseen garments using the same weight ${{\bm w}}$ learned on our original dataset. 
We also noticed that there exist some decorations such as pockets or shoulder boards on those garments, however, our method is robust enough to ignore these subtler features.

\section{Online Model Registration for Regrasping and unfolding}
\label{ch:unfolding}

As a part of the pipeline as shown in Fig. \ref{fig:entire_pipeline}, to unfold a garment, we can use the simulated models in the database to guide real object manipulation by registration.
In this pipeline, the registration results can be used for detection of regrasping points.
One of such scenarios is that unfolding a garment by iterative registration between the reconstructed mesh model and the database mesh model, and then regrasping. 
After several steps of regrasping, the robot holds the garment at two desired positions. 
Using a long-sleeve shirt as an example, we defined the optimal grasping positions on the two sleeves, respectively. 
The regrasping is built on the recognition pipeline described in the previous section. Once we have a recognized 3D object model from the database, we can perform a registration-search that looks for an optimal registration between the model and physical garment over the entire mesh model.  Once registered, we can then predict the best regrasping point in 3-dimensional space and guide the other hand to approach and regrasp at this point.  We do this using a fast, two-stage deformable object registration algorithm that integrates off-line simulated results with online localization and uses a novel non-rigid registration method to minimize energy differences between source and target models.
Then, we use a constrained weighted metric for evaluating grasping points during regrasping, which can also be used for a convergence criterion.

\begin{figure}[!htb]
\centering
\includegraphics[width=0.48\textwidth]{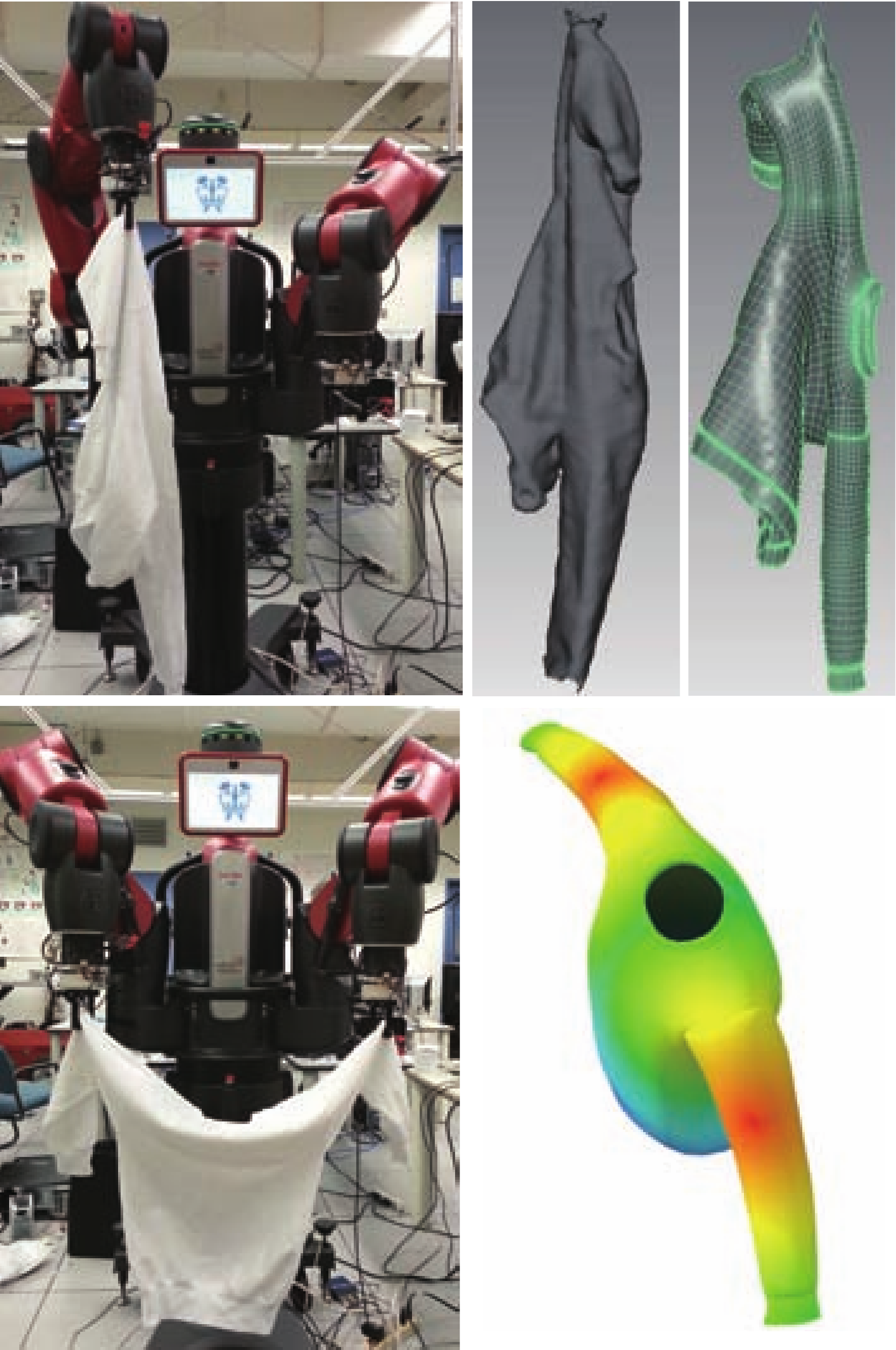}
\caption{
Our application scenario: a Baxter robot picks up a garment to recognize its pose via reconstruction. 
By deformable registration between the simulated mesh (top right) and reconstructed mesh (top middle), we obtain the regrasping point via a pre-determined point on the simulated mesh.
A long-sleeve shirt mesh with rendered weighted Gaussian distribution is shown on the bottom right. Red color indicates a higher score for evaluation of the grasping points, which are designated as the elbows of the sleeves.
The final unfolding result by the Baxter robot is shown on the bottom left.
}
\label{fig:scenario_regrasp}
\end{figure}

\begin{figure*}[!htb]
  \centering
  \includegraphics[width=0.99\textwidth]{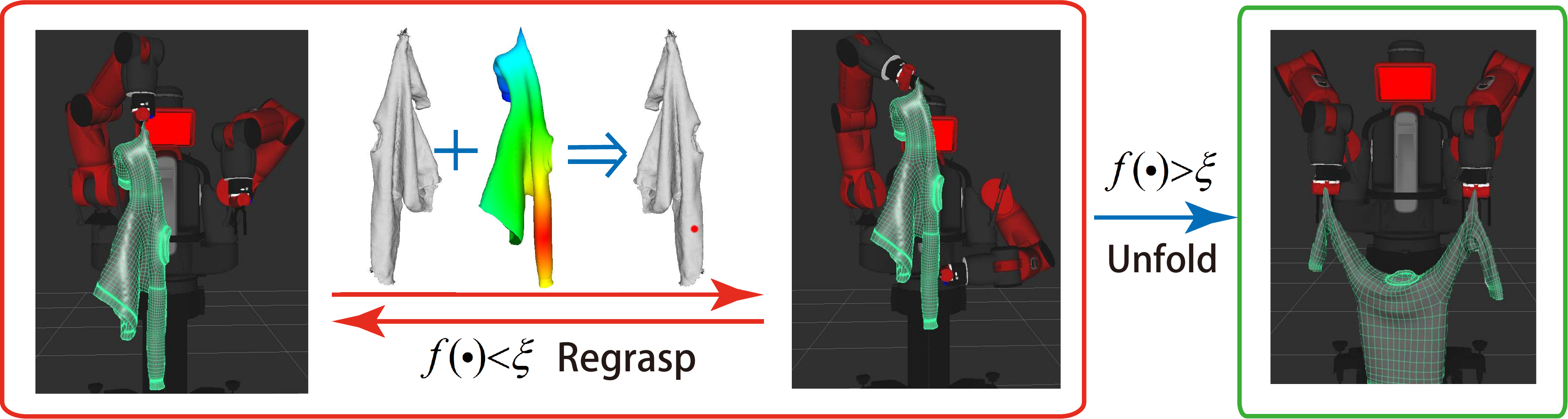}
  \caption{
		If the recognition is not successful or the pose is improper evaluated by the $f(\cdot)$ function, the robot will regrasp the object and repeat the step of pose estimation (the red rectangle). 
		Note that by registration between the reconstructed mesh from the Kinect and the simulate mesh from pose estimation, the robot knows where to regrasp subsequently as indicated by a red dot.
		This will be evaluated by the $f(\cdot)$ function. If $f(\cdot) < \xi$, the robot moves to unfold phase (the green rectangle). If this is not the case, the robot regrasps and goes back to pose estimation.}
\label{fig:unfolding_flowchart}
\end{figure*}

\subsection{Problem Formulation}
\label{section:unfolding_approach}
Our objective is to put the garment into a certain \emph{configuration}, which is defined as the relative grasping points on the garment~\cite{LiIROS2014}, such that the garment can be easily placed flat on a table for the folding process.
This problem can be formulated as a mathematical optimization problem:
\begin{equation}
    \max_{\mathbf{x}_L, \mathbf{x}_R} f(\mathbf{x}_L, \mathbf{x}_R).
    \label{eq:init}
\end{equation}
Here $\mathbf{x}_L, \mathbf{x}_R \in \mathbb{R}^2$ \footnote{Each garment mesh is defined in a $UV$ 2-dimensional parameter space. When we choose a grasping point, we choose a particular set of $UV$ parameters, which then will be mapped by registration with the sensed garment to a grasping point in $\mathbb{R}^3$.} are the positions of the left and right grasping points on the garment (the configuration) and the function $f$ is an evaluation function for such a configuration.
We seek a principled way to build a feedback loop for garment regrasping, which allows us to grasp at pre-determined points on the garment, and place it flat.

Suppose the candidate garment is a long-sleeve shirt which we want to unfold and place flat. 
A desired solution is grasping points $(\mathbf{x}_L^*, \mathbf{x}_R^*)$ lying on the elbows of the sleeves.
Our goal is to find a pair of grasping points $(\mathbf{x}_L, \mathbf{x}_R)$, through a series of regrasping procedures that will converge to a value close to $(\mathbf{x}_L^*, \mathbf{x}_R^*)$.   
We need a quantitative function defined on the \emph{pose} of the garment (i.e., where the robot arm grasps the garment) in order to evaluate how good a grasping point is.
While this can be computed on the continuous surface of the garment, we can also discretize the garment into a set of {\emph{anchor points}} ${{S}_g}$, which typically contains about $15$ points for a garment in our database.
After such quantization, the garment pose recognition can be treated as a discrete classification problem, which the current robotics system is able to handle reliably.
This also simplifies the definition of the objective function, which then becomes a $2$D score table or a matrix, given our robot has two arms.

\begin{figure}[t]
  \center
  \begin{tabular}{cc}
 \includegraphics[width=0.25\textwidth]{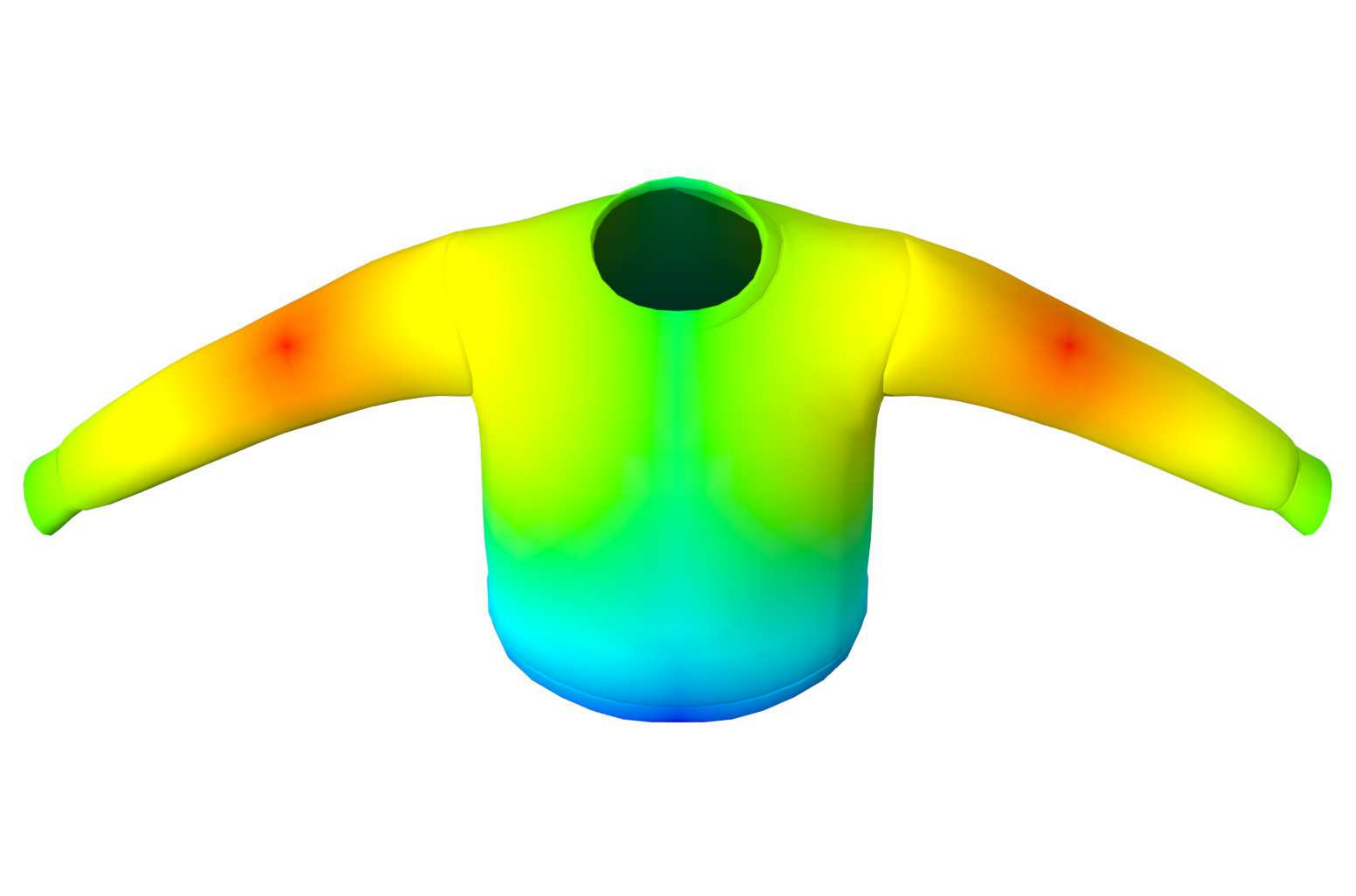} &
 \includegraphics[width=0.1\textwidth]{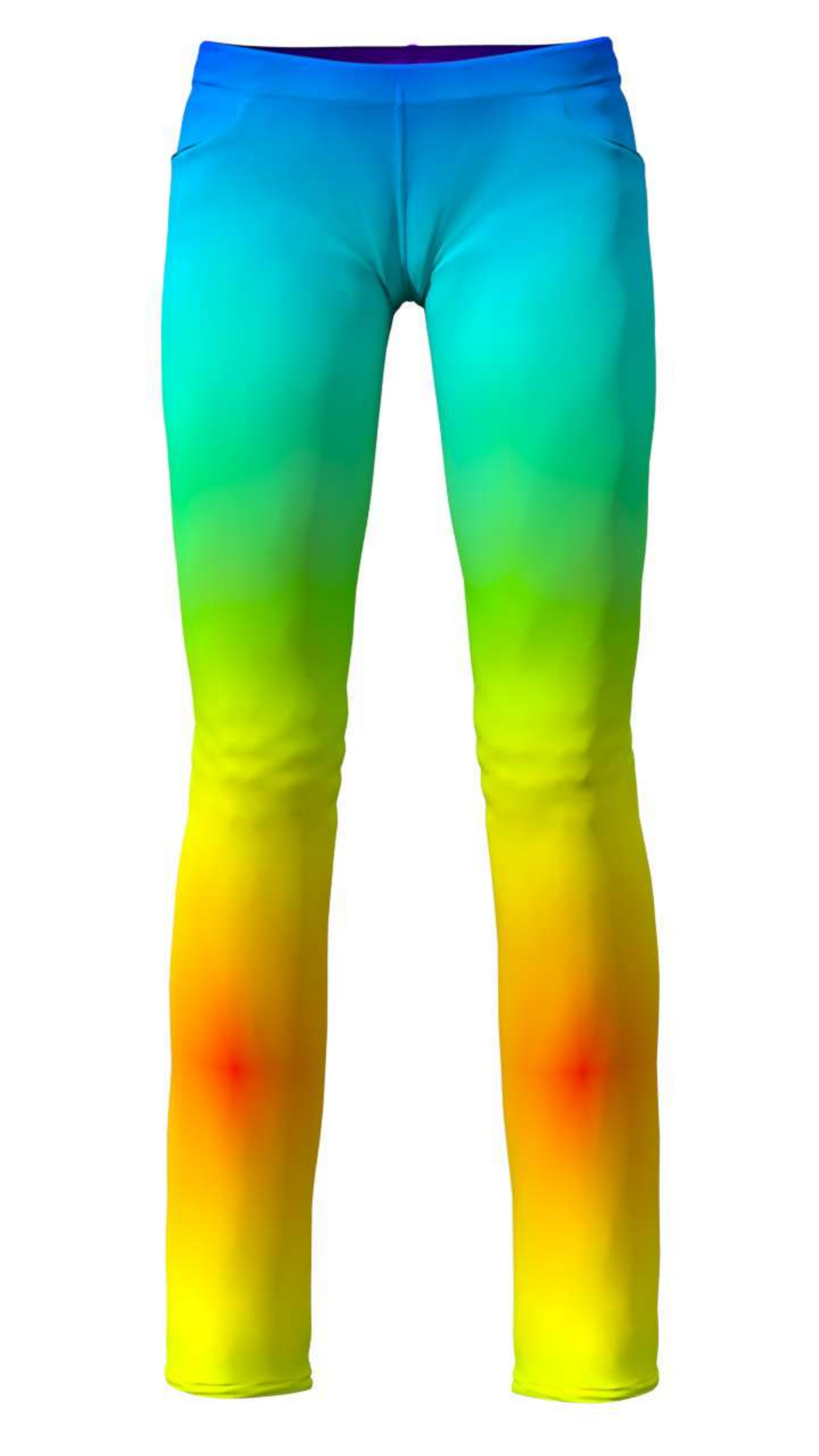} \\
		(a) & (b) \\
  \end{tabular}
  \caption{Visualization of the defined objective used in this section.
    (a) A long-sleeve shirt model rendered with weighted Gaussian distribution. (b) A pants model rendered with weighted Gaussian distribution.
    When a point is given over the garment surface, we can then evaluate the score by the objective function $f(\cdot)$.
  }
  \label{fig:objective}
\end{figure}

Details of the optimization procedure and inference can be found in~\cite{LiICRA2015}.
The objective function which needs to be maximized finally can be written as:
\begin{equation}
    \begin{aligned}[rl]
        \ln f({\mathbf{x}}_L,{\mathbf{x}}_R) = 
        & -\sum_{\mathbf{x}_l, \mathbf{x}_r \in S_g } \Big(
        \sigma_l \lVert \mathbf{x}_l - \mathbf{x}_L^* \rVert^2  \\
        & + \sigma_r \lVert \mathbf{x}_r - \mathbf{x}_R^* \rVert^2
        - \ln p(\mathbf{x}_l, \mathbf{x}_r|y) \Big). 
    \end{aligned}
    \label{eq:finalobj}
\end{equation}
The related parameters in the objective such as $\sigma_L$ and $\sigma_R$ are set depending on the desired configuration.
For example, for long-sleeve shirts, we set $\mathbf{x}_L^*$ and $\mathbf{x}_R^*$ on the elbow of the two sleeves.
The Gaussian formulation ensures a smooth decrease from the expected grasping points, as visualized in Figure~\ref{fig:objective} as an example.


\subsection{Deformable Registration}
\label{sec:registration}

After obtaining the location of the current grasp point, we seek to
register the reconstructed 3D model to the ground truth garment mesh
to establish point correspondences. The input to the registration is a
canonical reference (``source'') triangle mesh $S$ that has been
computed in advance and stored in the garment database, and a target
triangle mesh $T$ representing the geometry of the garment grasped by the
robot, as acquired by 3D scans of the grasped garment.

The registration proceeds in three steps. First, we scale the source
mesh $S$ to match its size to the target mesh $T$.  Next, we apply an
iterative closest point (ICP) technique to rigidly transform the
source mesh $S$ (i.e., via only translation and rotation). Finally, we
apply a non-rigid registration technique to locally deform the source
mesh $S$ toward the target $T$.

\paragraph{Scaling}
First, we compute a representative size for each of the source and
target meshes. For a given mesh, let $a_i$ and $\mathbf{g}_i$ be the
area and barycenter of the $i$th triangle. Then the area-weighted center
$\mathbf{c}$ of the mesh is
\begin{equation}
\mathbf{c} = \sum_i^{N_S} a_i \mathbf{g}_i \Big / \sum_i^{N_S} a_i,
\end{equation}
where $N_S$ is the number of vertices of the source mesh $S$.
Given the area-weighted center, the representative size $l$ of the
mesh is given by
\begin{equation}
l = \sum_i^{N_S} a_i \|\mathbf{g}_i - \mathbf{c}\| \Big / \sum_i^{N_S} a_i.
\end{equation}
Let the representative sizes of the source and target meshes be $l_S$ and $l_T$, respectively.
Then, we scale the source mesh by a factor of $l_T/l_S$.

\paragraph{Computing the rigid transformation}
We use a variant of ICP~\cite{Besl:1992} to compute the rigid
transformation. ICP iteratively updates a rigid transformation by
(a) finding the closest point $\mathbf{w}_j$ on the target mesh $T$
for each of the vertices $\mathbf{v}_j$ of the source mesh $S$, (b)
computing the optimal rigid motion (rotation and translation) that
minimizes the distance between $\mathbf{w}_j$ and $\mathbf{v}_j$, and
then (c) updating the vertices $\mathbf{v}_j$ via this rigid motion.

\begin{figure}[!htpb]
  \center
  \begin{tabular}{c}
 \includegraphics[width=0.35\textwidth]{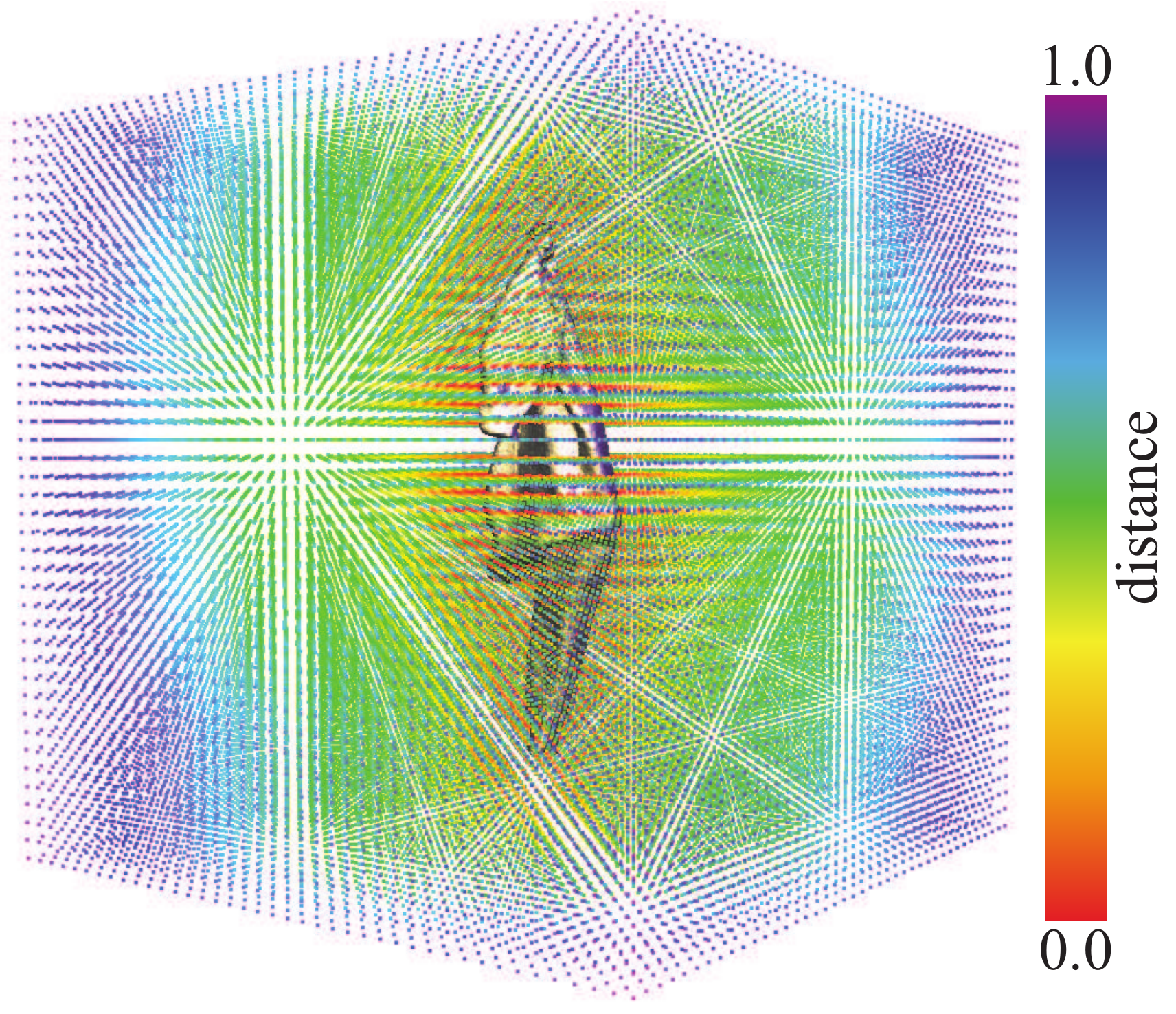}
  \end{tabular}
  \caption{Visualization of distance function given a mesh. 
	The color bar on the right shows the normalization distance.
  }
  \label{fig:distfunc}
\end{figure}

To accelerate the closest point query, we prepare a grid data
structure during preprocessing. For each grid point, we compute the
closest point on the target mesh using fast sweeping~\cite{Tsai:2002},
and store for runtime using both the found point and its distance
to the grid point as shown in Figure~\ref{fig:distfunc}. 
At runtime, we approximate the closest point query for
vertex $\mathbf{v}_j$ by searching only among those eight precomputed
closest points corresponding to the eight grid points surrounding
$\mathbf{v}_j$, thereby reducing the complexity of the closest point
query to $O(1)$ per vertex.

After establishing point correspondences, we compute the optimal
rotation and translation for registering $\mathbf{v}_j$ with
$\mathbf{w}_j$~\cite{Besl:1992}.
We iteratively compute point correspondences and rigid motions until
successive iterations converge to a fixed rigid motion, yielding a
rigidly registrated source mesh $\bar{S}$.

\paragraph{Non-rigid registration}
Given a candidate source mesh $\bar{S}$ obtained via rigid
registration, our non-rigid registration seeks the vertex positions
$\mathbf{v}_j$ of the source mesh $S$ that minimize 
\begin{equation}
\label{eqn:energy}
E_{\bar{S},T}(S) = E_{\textrm{fit}}(S,T) + E_{\textrm{def}}(S,\bar{S}),
\end{equation}
where $E_{\textrm{fit}}(S,T)$ penalizes discrepancies between the source
and target meshes, and $E_{\textrm{def}}(S,\bar{S})$ seeks to limit and
regularize the deformation of the source mesh away from its rigidly
registrated counterpart $\bar{S}$. The term
\begin{equation}
E_{\textrm{fit}} = \sum_{i=1}^{N_S} \left( \mbox{dist}(\mathbf{g}_i) \right)^2 \bar{A}_i,
\end{equation}
penalizes deviation of the source and target meshes.  Here
$\mathbf{g}_i$ is the barycenter of the triangle $i$, and
$\mbox{dist}(\mathbf{g}_i)$ is the distance from $\mathbf{g}_i$ to the
closest point on the target mesh. As in the rigid case, we use the
precomputed distance field to query for the distance. 

It might appear that the fitting energy $E_{\textrm{fit}}$ could be
trivially minimized by moving each vertex of mesh $S$ to lie on mesh
$T$. In practice, however, this does not work
because all of the geometry of the precomputed reference mesh
$\bar{S}$ is discarded; instead, the geometry of this mesh, which was
precomputed using fabric simulation, should serve as a prior. Thus, we
introduce a second term to retain as much as possible the geometry of
the reference mesh $\bar{S}$:

The deformation term $E_{\textrm{def}}(S,\bar{S})$, derived from a physically
based energy (e.g., see \cite{Grinspun:2003}), is a sum of three terms
\begin{equation}
E_{\textrm{def}}(S,\bar{S}) = \kappa E_{area} + \beta E_{angle} + \alpha E_{hinge},
\end{equation}
where $\alpha$, $\beta$ and $\kappa$ are user-specified
coefficients.The term
\begin{equation}
E_{area} = \sum_{i=1}^{N_S} \frac{1}{2} \left( \frac{A_i}{\bar{A}_{i}} - 1 \right )^2 \bar{A}_i,
\end{equation}
penalizes changes to the area of each mesh triangle. Here $A_i$ is the
area of the triangle $i$, and $\bar{\cdot}$ refers to a corresponding
quantity form the undeformed mesh $\bar{S}$. The term 
\begin{equation}
E_{angle} = \sum_{i=1}^{N_S} \sum_{k=1}^3 \frac{1}{6} \left( \frac{\theta_{ik}}{\bar{\theta}_{ik}} - 1 \right )^2 \bar{A}_i,
\end{equation}
penalizes shearing of each mesh triangle, where $\theta_{ik}$ is the
$k$th angle of the triangle $i$. The term $E_{hinge}$~\cite{Grinspun:2003} 
\begin{equation}
E_{hinge} = \sum_e (\theta_e - \bar{\theta}_e)^2 \|\bar{e}\| / \bar{h}_e,
\end{equation}
penalizes bending, measured by the angle formed by adjacent triangles.
Here $\theta_e$ is the \emph{hinge angle} of edge $e$, i.e., the angle
formed by the normals of the two triangles incident to $e$;
$\|\bar{e}\|$ is the length of the edge $e$, and $\bar{h}_e$ is a
third of the sum of the heights of the two triangles incident to the
edge $e$.

We used the secant-version of the L-M method\cite{Madsen:2004} to seek
the source mesh $S$ that minimizes the energy Eq.(\ref{eqn:energy}). 
Sample registration results are shown in Figure~\ref{fig:registration}.

\subsection{Grasping Point Localization}
\label{grasp_pt_location}
We use a pre-determined anchor point (e.g., elbow on the sleeve of a long-sleeve shirt) to indicate a possible regrasping point.
The detection of the regrasping point can be summarized in two steps: \emph{global localization} and \emph{local refinement}.
\emph{Global localization} is achieved by deformable registration. 
The registered simulation mesh will provide a 3D regrasping point from the recognized state which will be then mapped onto the reconstructed mesh.
Details of \emph{local refinement} can be found in~\cite{LiICRA2015}.

In order to improve the regrasping success rate, we propose a step of \emph{local refinement}.
The point on the actual garment may be hard to grasp for several reasons. 
One is that during the garment manipulation steps, such as rotation, the curvature over the garment may change. 
Another reason is that when considering the width of robot hand gripper, a ridge curve with proper orientation and width should be selected for regrasping.
We consider the proper orientation as a direction perpendicular to the opening of the gripper.
Therefore, we propose an efficient 1D blob curvature detection algorithm that can find a refined position in the local area over the garment surface via an IR range sensor.

In our experiment, the Baxter robot is equipped with a IR range sensor close to the gripper as shown in \Fig{ir_sensor} top. 
Once the gripper moves to the same height of the predicted 3D regrasping point from registration, it will perform a horizontal scan search to achieve a refinement of the local grasping point, moving from one side to the other, so that the IR sensor will scan over the full local curvature.

We then apply a curvature detection algorithm that convolves the IR depth signal with a fixed width kernel, where the width is determined by the opening of the gripper.
Here we use a Laplacian-Gaussian Kernel $g''(x)$: 
\begin{equation}
\label{Laplacian_Gaussian}
g''(x) = (\frac{{{x^2}}}{{{\sigma ^4}}} - \frac{1}{{{\sigma ^2}}}){e^{ - \frac{{{x^2}}}{{2{\sigma ^2}}}}}
\end{equation}
where $x$ is the depth signal, and $\sigma$ is the width parameter.

\subsection{Convergence}

After the regrasping is finished, we evaluate the current grasping configuration by the objective function $f(\cdot)$.
If $f(\cdot)$ is greater than a given value $\xi$, which means the grasping points are on the desired positions, and the robot will then stop regrasping and enter the placing flat mode.
The two arms will open to slightly stretch the garment and place it on a table. 
The overall algorithm is summarized in Algorithm~\ref{alg:regrasping}.

\begin{algorithm}[!htpb]
  \caption{Iterative Procedure for Regrasping}
  \label{alg:regrasping}
  \KwIn{\
		Simulation meshes = $M = {\{M_{sim}^{1}, M_{sim}^{2},...,M_{sim}^{n}\}}$\;
		Trained classifier $C$\;
		Objective function $f$\;}
  \KwOut{\
	Two desired grasping points $X_L$ and $X_R$\;}
	${f_{score}} \leftarrow \infty$, $i \leftarrow1$\;
	\While{${f_{score}} < \xi$}{
	Pick up at a grasping point $P_{grasp}^i$\;
	$M_{rec}^{i} \leftarrow$ 3D Reconstruction\;
	$M_{sim}^{i} \leftarrow C(M_{rec}^{i};M_{sim}^{1}, M_{sim}^{2},...,M_{sim}^{n})$ //Recognition\;
	$P_{grasp}^{i+1} \leftarrow$ Reg($M_{rec}^{i}$, $M_{sim}^{i}$) //Registration\;
	$f_{score} \leftarrow f({P_{grasp}^i}, P_{grasp}^{i+1})$\;
	$i \leftarrow i+1$\;
	}
	\Return{$\mathbf{x}_L$ and $\mathbf{x}_R$}\;
	Place the garment flat on a table\;
\end{algorithm}

\subsection{Experimental Results}
\label{section:unfolding_experiments}

To evaluate our results, we tested our method on several different garments such as long-sleeve shirts and pants for multiple trials.

\begin{figure}[!htpb]
\center
\begin{tabular}{c}
 \includegraphics[width=0.48\textwidth]{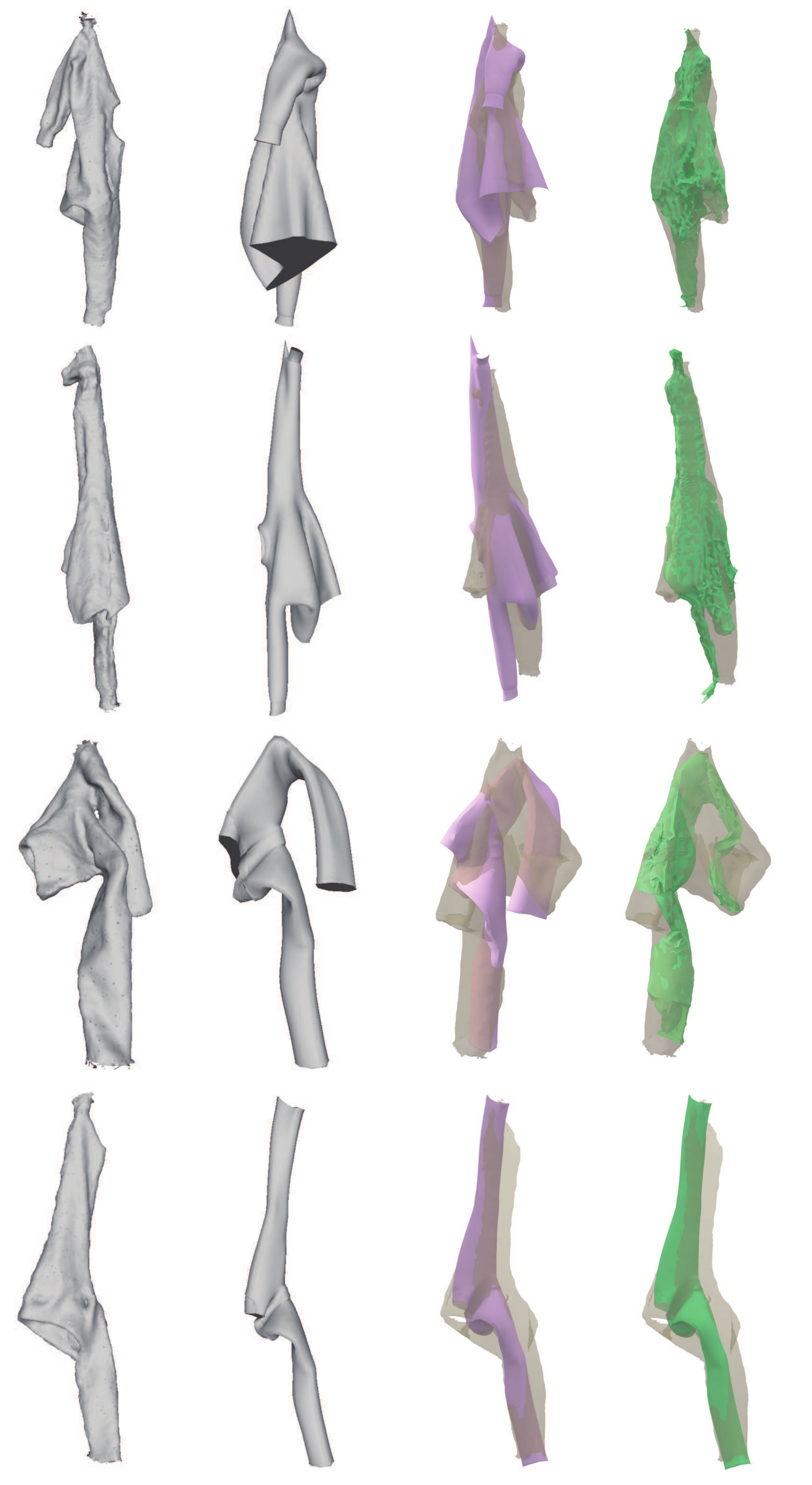} \\
\end{tabular}
\caption{Registration examples. 
	{{\sc{First Row:}} A long-sleeve shirt grasped at elbow.} 
	{{\sc{Second Row:}} A long-sleeve shirt grasped at sleeve end.}
	{{\sc{Third Row:}}A pair of pants grasped near knee.} 
		{{\sc{Fourth Row:}} A pair of pants grasped near ankle.} 
		Each row depicts from left to right: a reconstructed mesh, the predicted mesh from the database, rigid registration only, and rigid plus non-rigid registration.
}
\label{fig:registration}
\end{figure}

\begin{figure*}[!htpb]
  \centering
  \includegraphics[width=0.99\textwidth]{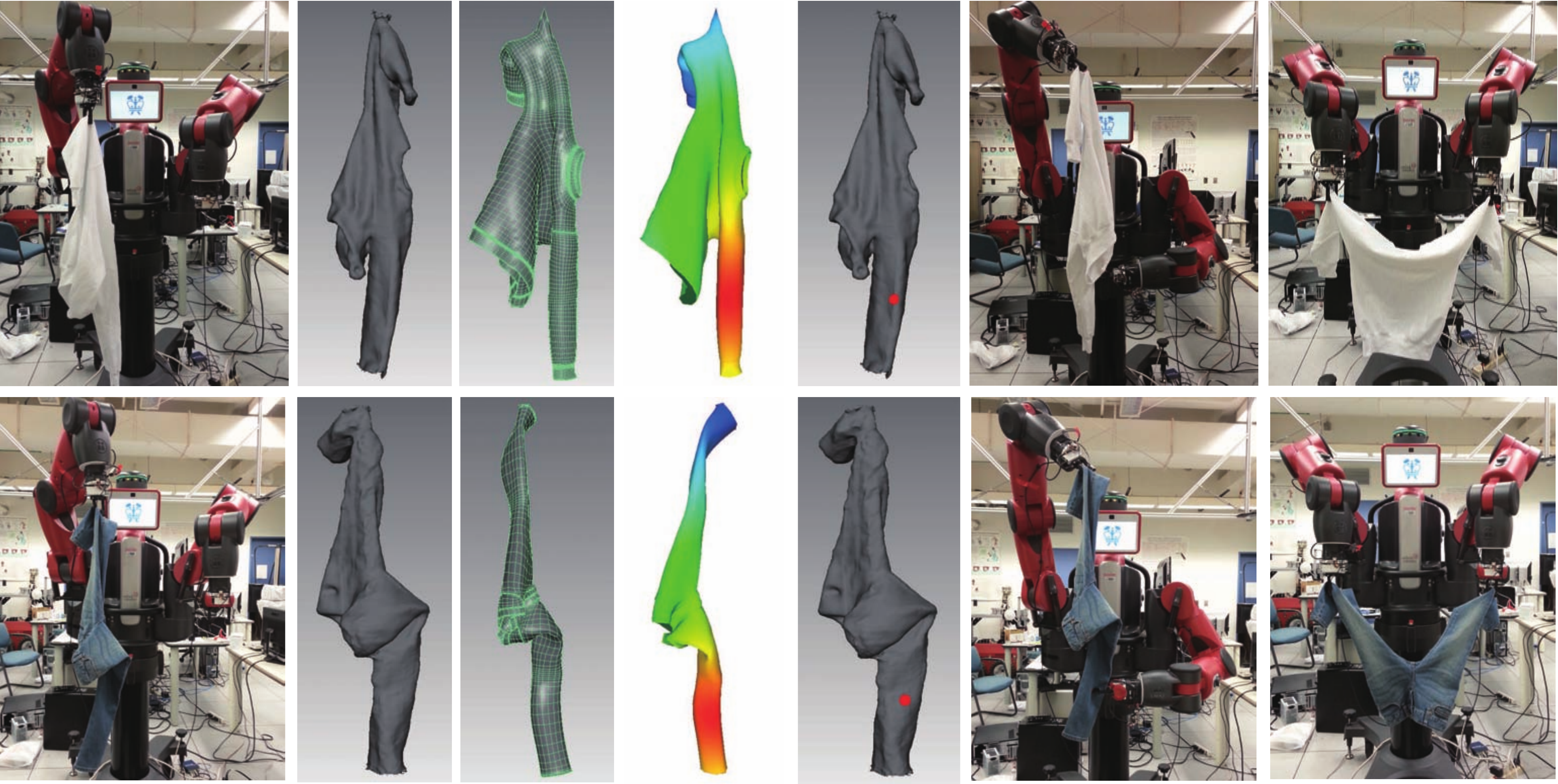}
  \caption{ Examples of each step in our unfolding procedure.
	For each row from left to right is: a snapshot of initial pick up, a 3D reconstructed mesh, a predicted mesh from database, the predicted mesh with weighted Gaussian distribution distance, predicted regrasping point on the 3D reconstructed mesh, a snapshot of regrasping, and finally a snapshot of unfolding. 
    {\sc Top Row: } The Baxter robot unfolds a long-sleeve shirt following pick up. 
    {\sc Bottom Row: }  The Baxter robot unfolds a pair of pants following pick up.}
\label{fig:iter_regrasp}
\end{figure*}

Below, we briefly recap the pose recognition method. Details can be found in the previous section.
We first pick up the garment at a random point.
In the online recognition stage, we use a Kinect sensor to capture depth images of different views of the garment while it is being rotated by a robotic arm.
The garment is rotated $360^\circ$ clockwise and then $360^\circ$ counter-clockwise to obtain about $550$ depth images for an accurate reconstruction.
We reconstruct a 3D mesh model from the depth image segmentation and volumetric fusion. 
Then with an efficient 3D feature extraction algorithm, we build up a binary feature vector and finally match against the offline database for pose recognition.
One of the outputs is a high-quality reconstructed mesh, which is used for 3D registration and accurate regrasping point prediction, as described below.

\subsection{Registration}
We apply both rigid and non-rigid registrations. 
The rigid registration step mainly focuses on mesh rescaling and alignment, whereas the non-rigid registration step refines the results and improves the mapping accuracy.
In Figure~\ref{fig:registration}, we compare the difference between using rigid registration only and using rigid plus non-rigid registration side by side. 
We can clearly see that with non-rigid registration, the two meshes are registered more accurately. 
In addition, the location of the designated grasping points on the sleeves are also closer to the ground truth points. 
Note that for the fourth row, after the alignment by the rigid registration algorithm, the state is evaluated as a local minimum.
Therefore, there is no improvement by the following non-rigid registration. 
But as we can see from the visualization, such a case is still good enough for finding point correspondence.

\begin{table*}
\center
\begin{tabular}{c|p{2cm}|p{2cm}|p{2cm} | p{2cm}} \hline
	Source Mesh &  \centering S to T (R) & \centering T to S (R) & \centering S to T (R+N) &\centering T to S (R+N)  \tabularnewline \hline
	Long-Sleeve T-Shirt 1 & \centering $0.0251$  & \centering $0.0188$  & \centering $0.0073$ & \centering $0.0129$ \tabularnewline
	Long-Sleeve T-Shirt 2 &  \centering  $0.0171$ & \centering $0.0218$ & \centering $0.0151$ & \centering $0.0213$ \tabularnewline
	Long Pants 1 &  \centering $0.0099$  & \centering  $0.0652$ & \centering $0.0060$ & \centering $0.0648$ \tabularnewline
	Long Pants 2 &   \centering  $0.0164$ & \centering  $0.0221$ & \centering $0.0044$ & \centering $0.0172$ \tabularnewline
	Shorts &  \centering $0.0238$  & \centering  $0.0149$ & \centering $0.0081$ & \centering $0.0092$ \tabularnewline \hline
	{\bf{Average}} & \centering ${\bf{0.0185}}$ & \centering ${\bf{0.0256}}$ & \centering $\bf{0.0070}$ & \centering $\bf{0.0250}$ \tabularnewline
	\hline
\end{tabular}
\caption{Registration results. We compare the source mesh (S) registered to the target mesh(T), and vice versa, for both rigid-only registration(R) and rigid plus non-rigid registration(R+N).
We can see that when the source mesh registered to the target mesh, the average error distance is less than the target mesh registered to the source mesh. This is because the source mesh is with less resolution, whose deformation can be easily computed to reach a minimum. Also, we can see that with additional non-rigid registration, the average error distance is reduced.}
\label{tbl:registration_compare}
\end{table*}

We also evaluate the registration algorithm on the entire database, which contains two stage, rigid registration using ICP algorithm and non-rigid registration algorithm.
To show the performance of our registration algorithm, the registration pairs are established with the knowledge that the recognition of the pose is $100$\% correct.
This will enable the registration to happen between the closest grasping location. 
Meanwhile, we design the registration experiments in two directions, the source mesh to the target mesh, and vice versa.
We also compare the registration results of the rigid registration, and the rigid plus the non-rigid registration for all the pairs.
Detailed results are shown in Table~\ref{tbl:registration_compare}.
For example, for the S to T(R), we first subdivides the source mesh into a set of disjoint triangulated patches, and generates a single sample point in each patch. 
Each sample point is also assigned the area of the patch it belongs to. 
Then, from each such sample point, we find the closest point on the target mesh, and sum up the distance of all point pairs and multiplied by the corresponding patch area.
Finally, the summed value is divided by the total area of the source mesh.

\subsection{Search for Best Grasping Point by Local Curvature}
\begin{figure}[!htpb]
\center
\begin{tabular}{c}
	\includegraphics[width=0.3\textwidth]{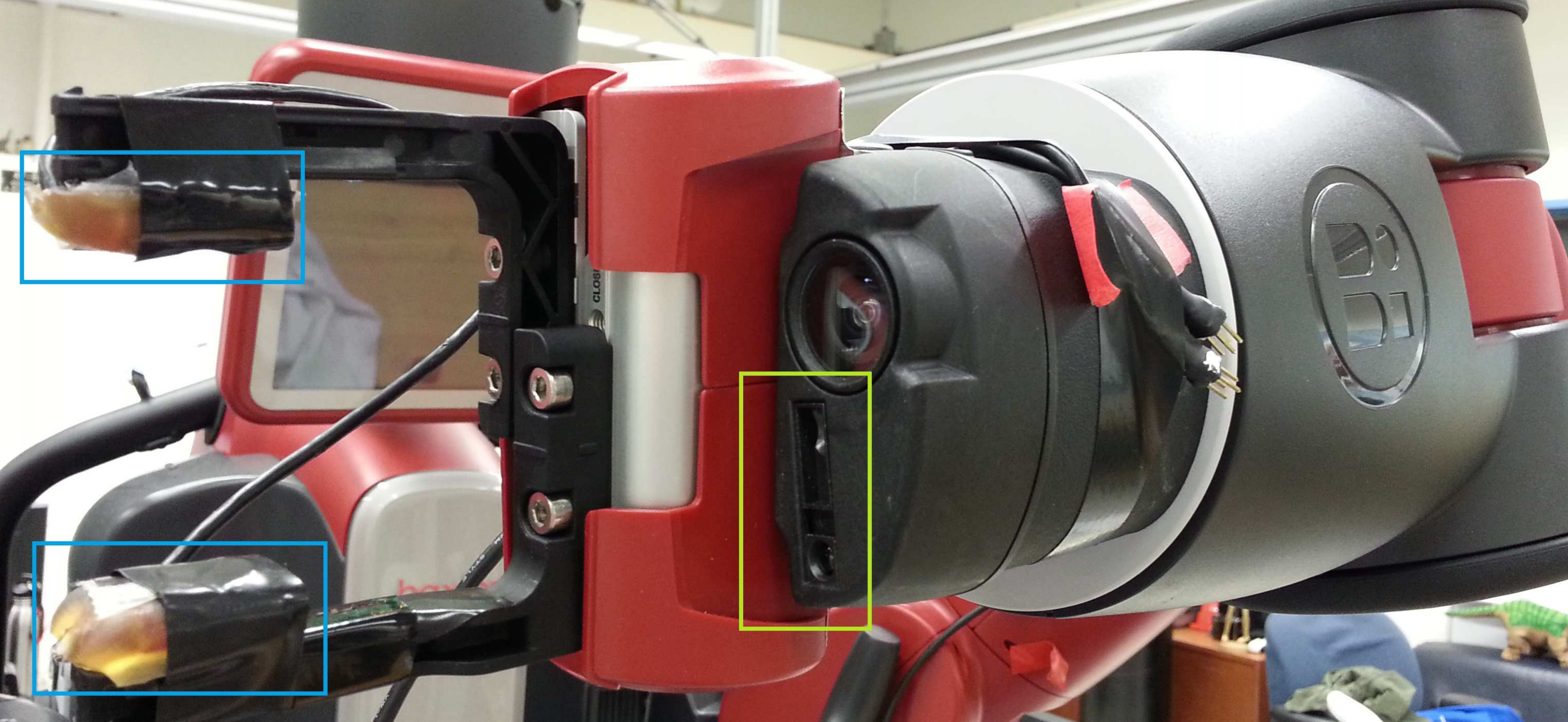} 
\end{tabular}
\begin{tabular}{cc}
	\includegraphics[width=0.2\textwidth]{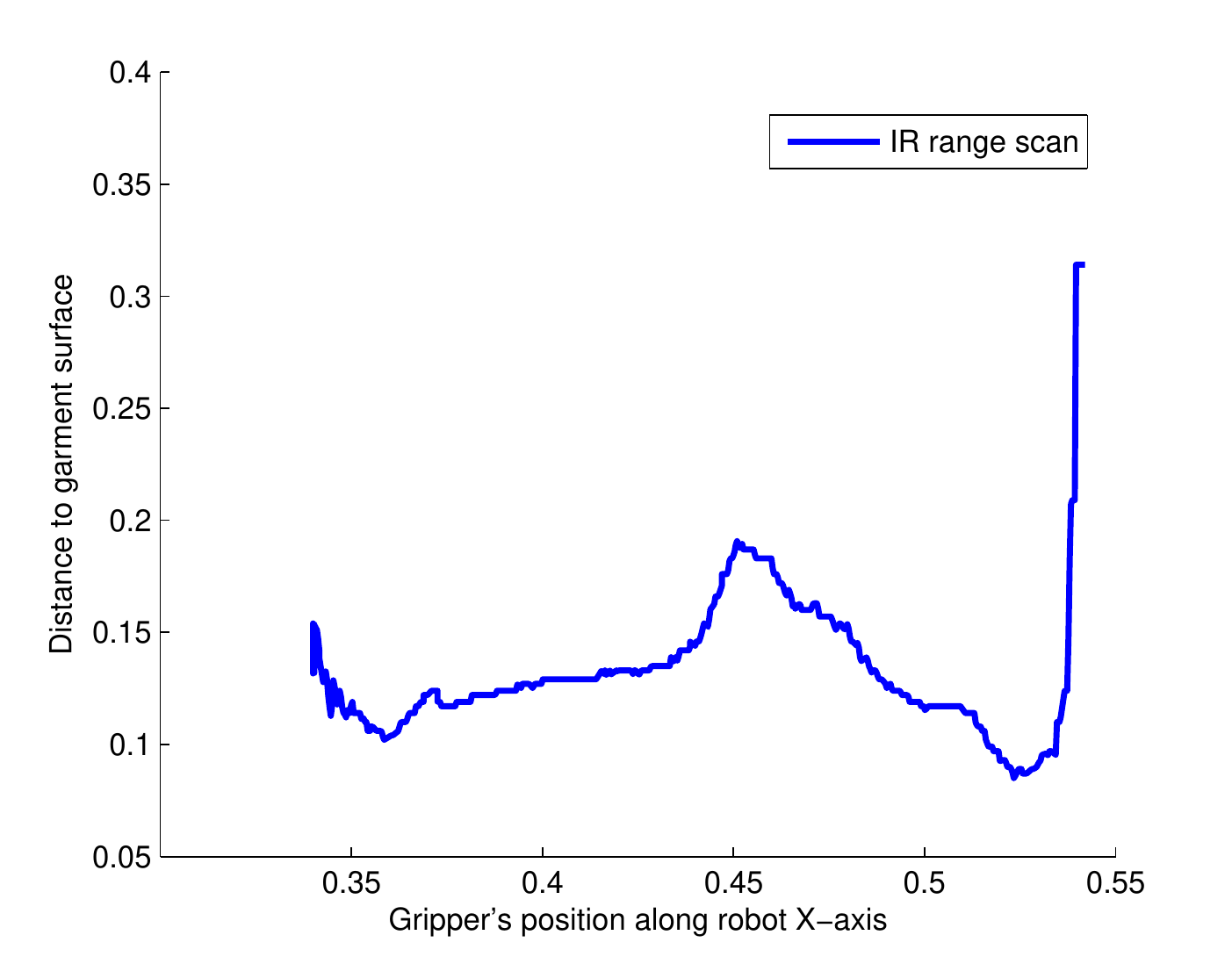}
	\includegraphics[width=0.2\textwidth]{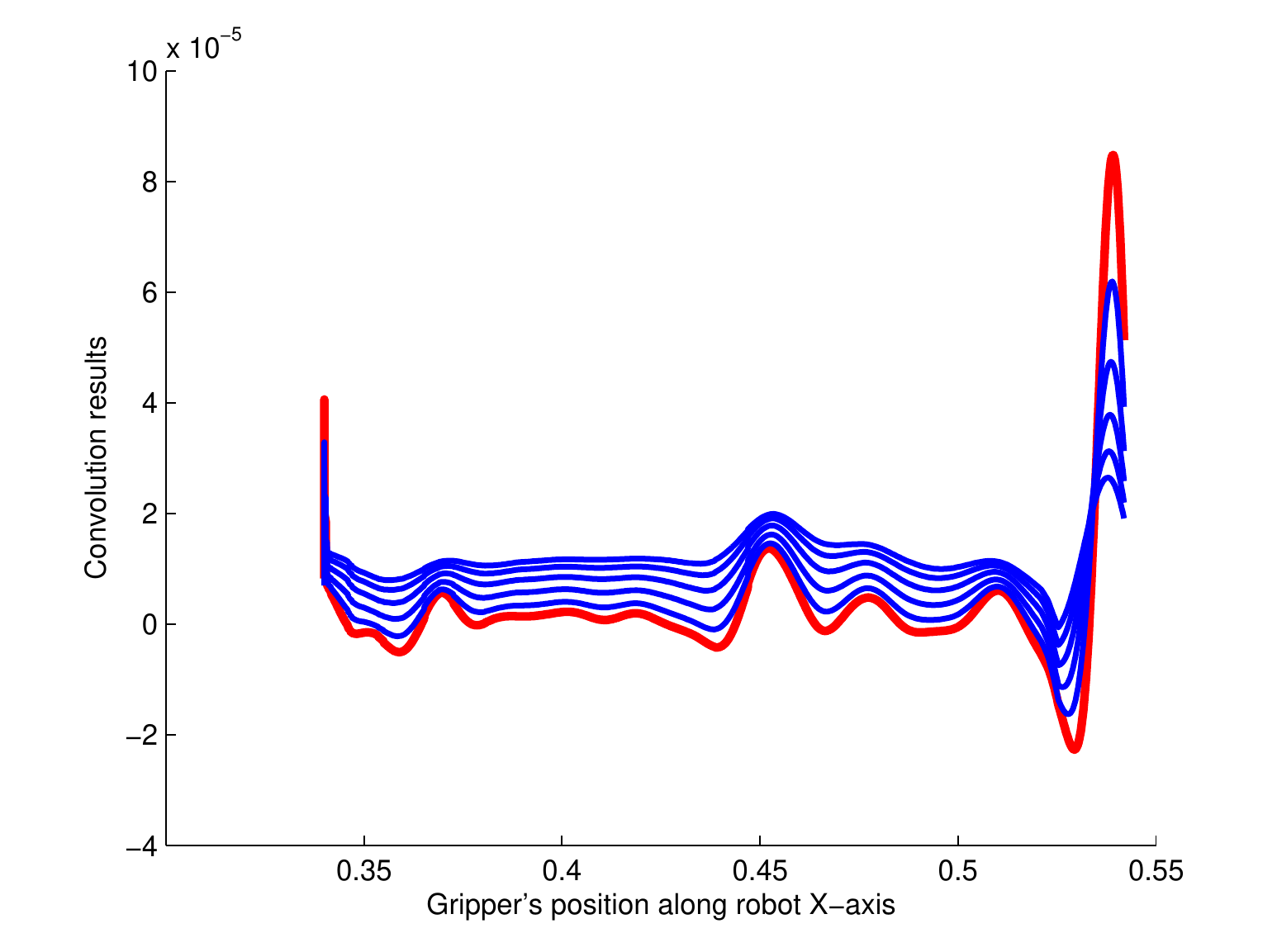}
\end{tabular}
\caption{IR range sensor scan example.
	{\sc{Top:}} An image of the Baxter hand. The IR range sensor is shown in yellow rectangle and two Tactile sensors in blue rectangles.
	{\sc{Bottom Left:}} Single reading plot from the IR range sensor.
	{\sc{Bottom Right:}} Convoluted result of the sensor reading and a Laplacian-Gaussian kernel with different kernel size. The lowest point (in red) is the place the gripper should grasp.
}
\label{fig:ir_sensor}
\end{figure}

Once we choose a potential grasping point, we can perform a search to find the best local grasping point for the gripper.
We are trying to a find a fold in the vicinity of the potential grasping point with a high local curvature tuned to the gripper width that allows for a stable grasp.
The opening size of the gripper is approximately $8cm$ and empirically we set $\sigma = 10$ in the equation~\ref{Laplacian_Gaussian}.
\Fig{ir_sensor} top shows a picture of the IR range sensor on the gripper.
A plot of its signal, as well as the convoluted signal, are shown in~\Fig{ir_sensor} bottom left and right.
We can clearly see that the response from the filter is at a minimum where the grasping should take place.
The tactile sensors then assure that the gripper has properly closed on the fabric.

\subsection{Iterative regrasping}
Figure~\ref{fig:iter_regrasp} shows two examples (long-sleeve shirt and pants) of iterative regrasping using the Baxter robot.
The robot first picks up a garment at a random grasping point.
Once the arm reaches a pre-defined position, the last joint of the arm starts to rotate and the Kinect will capture the depth images as it rotates, and reconstruct the 3D mesh in real-time.
After the rotation, a predicted pose is recognized~\cite{LiIROS2014} as shown in the third image of each row.
For each pose, we have a constrained weighted evaluation metric over the surface to identify the regrasping point as indicated in the fourth image.
By registration of the reconstructed mesh and predicted mesh from the database, we can map the desired regrasping point onto the reconstructed mesh.
The robot then regrasps by moving the other gripper towards it.
With our 1D blob curvature detection method, the gripper can move to the best curvature on the garment and regrasp, which increases the success rate.
The iterative regrasping stops when the two grasped points are the designated anchor points on the garment (e.g., elbows on the sleeves of a long-sleeve shirt).

\begin{figure*}[!htpb] 
  \begin{minipage}[]{0.4\textwidth} 
    \centering 
    \includegraphics[width=0.85\textwidth]{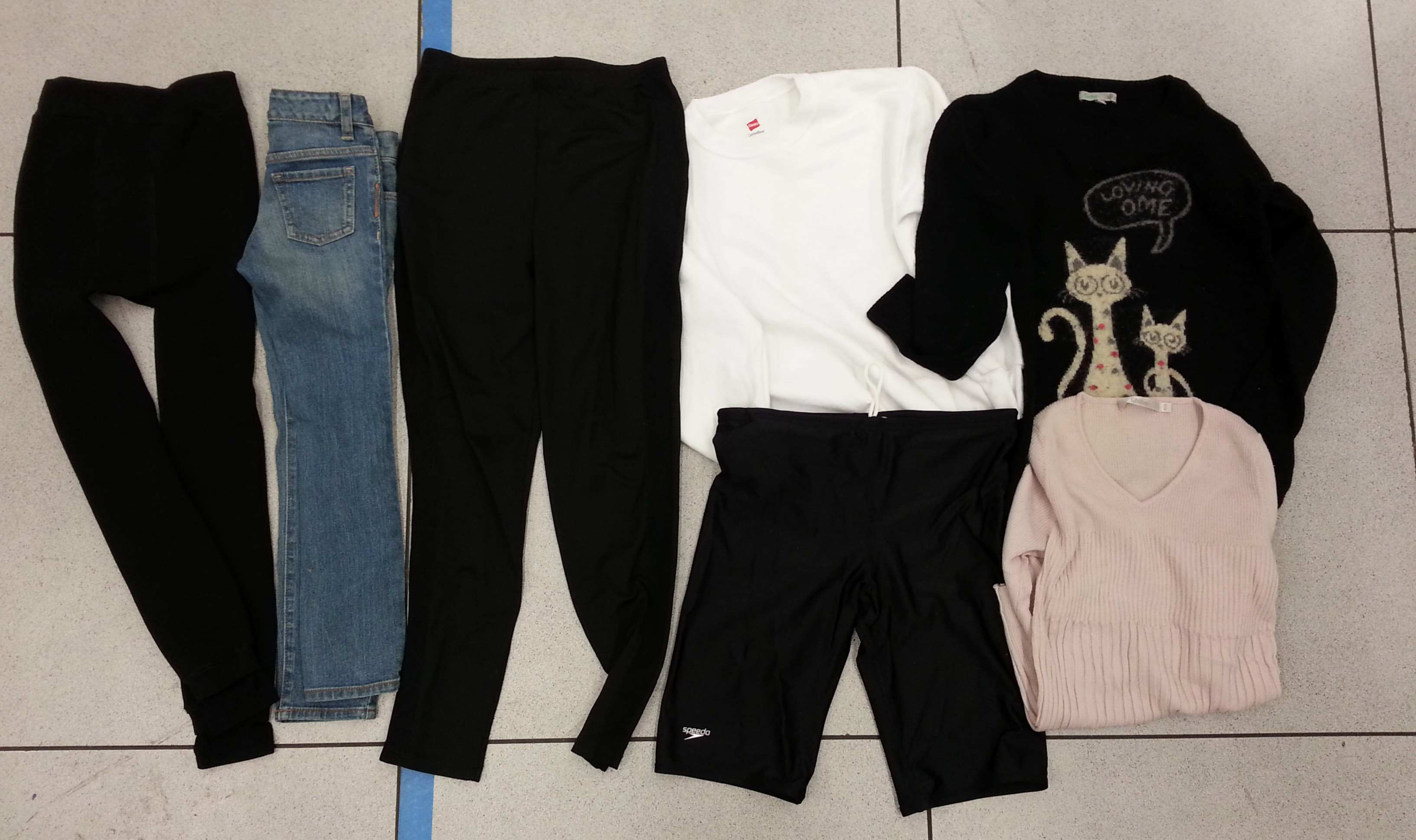} 
  \end{minipage}
  \begin{minipage}[]{0.6\textwidth} 
	\footnotesize
	\centering
    \begin{tabular}{c|p{0.6cm}|p{1.4cm}|p{1.3cm}|p{1.3cm}|p{1.8cm}} \hline
      Garment & \centering \# of Trial & \centering Successful Recognition & \centering Successful Regrasping & \centering Successful Unfolding & \centering Avg. \# of Regrasps Success Only  \tabularnewline \hline
      Sweatshirt & \centering $10$ & \centering $9/10$  & \centering $8/10$ & \centering $8/10$  & \centering $1.6$  \tabularnewline
			Sweater & \centering $10$ & \centering $8/10$  & \centering $8/10$ & \centering $7/10$  & \centering $1.6$ \tabularnewline
      Knitwear & \centering $10$ & \centering $9/10$  & \centering $9/10$ & \centering $8/10$  & \centering $1.7$ \tabularnewline
			Jeans & \centering $10$ & \centering $9/10$  & \centering $9/10$ & \centering $9/10$  & \centering $1.3$ \tabularnewline
			Pants & \centering $10$ & \centering $8/10$  & \centering $10/10$ & \centering $9/10$  & \centering $1.4$ \tabularnewline
			Leggings & \centering $10$ & \centering $8/10$  & \centering $9/10$ & \centering $8/10$  & \centering $1.4$ \tabularnewline
			Shorts & \centering $10$ & \centering $7/10$  & \centering $8/10$ & \centering $7/10$  & \centering $1.9$ \tabularnewline \hline
			{\bf{Average}} & \centering $10$ & \centering $8.3/10$  & \centering $8.7/10$ & \centering $8.0/10$  & \centering $1.6$ \tabularnewline
      \hline
    \end{tabular}
  \end{minipage} 
	\caption{{\sc{Left:} }A picture of our test garments.
	{\sc{Right:}} Results for each unfolding test on the garments. 
	We evaluate the results by recognition, regrasping, unfolding, and regrasping attempts for each test.
	The last row shows the average of each evaluation component.
	}
\label{fig:garment_stat_unfold}
\vspace{-0.5cm}
\end{figure*}

Figure~\ref{fig:garment_stat_unfold} left shows $7$ sample garments in our test, and the table on the right shows the results.
For each garment, we perform $10$ unfolding tests. 
We have on average an $83\%$ successful recognition rate for the pose of the objects over all the garments.
We have on average an $87\%$ successful regrasping rate for each garments, where regrasping is defined as a successful grasp of the other arm on the garment. 
$80\%$ of the time we are able to successfully unfold the garment, placing the grippers at the designated grasping points.
Unsuccessful unfolding occurred when either the gripper lost contact with the garment, or the gripper was unable to find a regrasping point.
Although we did not perform this experiment, it is possible to restart the method after one of the grippers loses contact as an error recovery procedure.

For the successful unfolding cases, we also report the average number of regrasping attempts.
The minimum number of regrasping attempts $=1$. This happens when the initial grasping is at one of the desired positions, and the regrasping succeeds at the other desired position (i.e., two elbows on the sleeves for a long-sleeve shirt). 
In most cases, we are able to successfully unfold the garments using $1-2$ regraspings.

Among all these garments, jeans, pants, and leggings achieve high success rate because of their unique layout when grasping at the leg position.
The shorts are difficult for both recognition and unfolding steps possibly because its ambiguous appearances in different grasping points.
One observation is that in a few cases, when the recognition is not accurate, our registration algorithm was sometimes able to find a desired regrasping point for unfolding.
This is an artifact of the geometry of pant-like garments where the designated regrasping points are at the extreme locations on the garments.

\section{Trajectory Optimization for Folding}
\label{ch:folding}

\begin{figure}[!htpb]
\begin{center}
  \centering
  \includegraphics[width=0.45\textwidth]{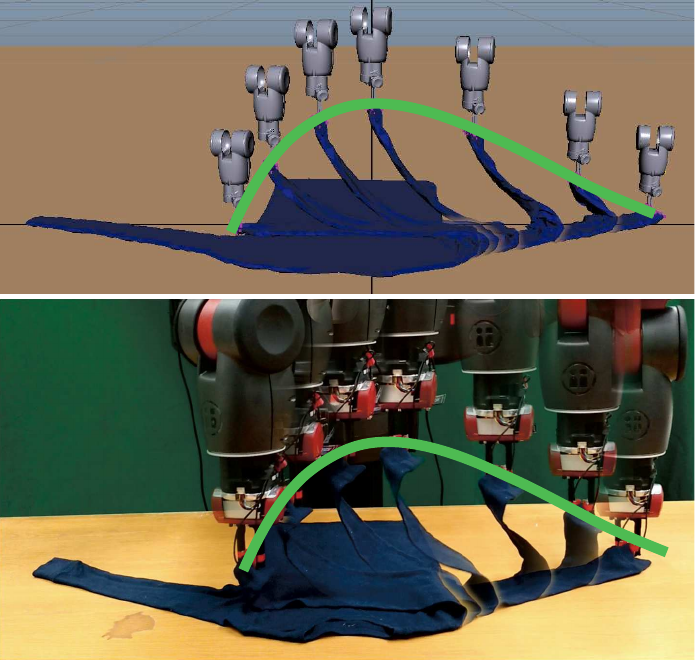}
\end{center}
   \caption{
	{\sc{Top:}} Comparison of our simulation of robotic manipulation. {\sc{Bottom:}} Real
robot implementation. The green curves show the virtual and the real trajectories for folding.}
\label{fig:intro}
\end{figure}

Robotic folding of a garment is a difficult task because it requires sequential manipulations of a highly unconstrained, deformable object.
Given the garment shape, the robot can fold it by following a folding plan~\cite{millerICRA2011}~\cite{Milleretal_IJRR2012}.
However, the layout of the same folding action can vary in terms of the material properties such as cloth hardness and the environment such as friction between the garment and the table.
Given the starting and ending folding positions, different folding trajectories will lead to different results.
In this section, we propose a novel method that learns optimal folding trajectory parameters from predicted thin shell simulations of similar garments, which can then be applied to a real garment folding task (see Figure~\ref{fig:intro}). 
We first present an online optimization algorithm that learns optimal trajectories for manipulation from mathematical model evolution combined with predictive thin shell simulation.
Meanwhile, a novel approach is introduced that can adjust the simulation environment to the robot working environment for the purpose of creating a similar manipulation result. 
Then, with the learned simulation results, we introduce a fast and robust algorithm that can detect garment key points such as sleeve ends, collar, and waist corner, automatically. These key points can be used for folding plan generation. 
The trajectories themselves are general in that they can be scaled to accommodate similar garments of different size.

\begin{figure}[!htpb]
\begin{center}
 \includegraphics[width=0.44\textwidth]{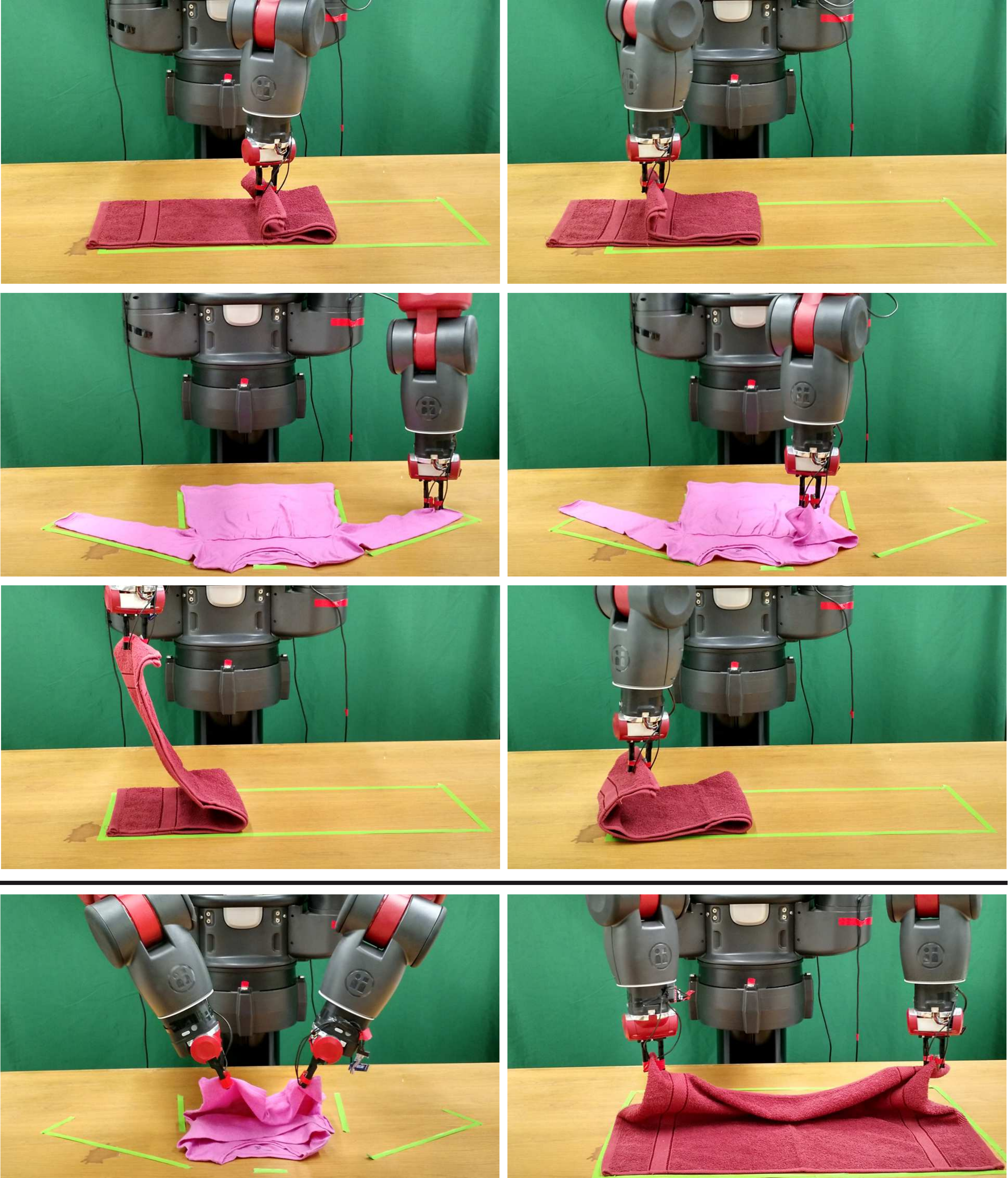}
\end{center}
   \caption{Failure example with improper folding trajectories.
	{\sc{First Row:}} Folding trajectory is low and flat that causes drift to the towel and long-sleeve T-Shirt.
	{\sc{Second and third Rows:}} Folding trajectory is too high when the gripper approaching the target folding position that piles up the towel.
	{\sc{Fourth Row:}} Dual-arm folding. If the distance between the two arms is too close, the folding may fail.}
\label{fig:failure_examples}
\end{figure}

\begin{figure*}[!htpb]
  \centering
  \includegraphics[width=0.99\textwidth]{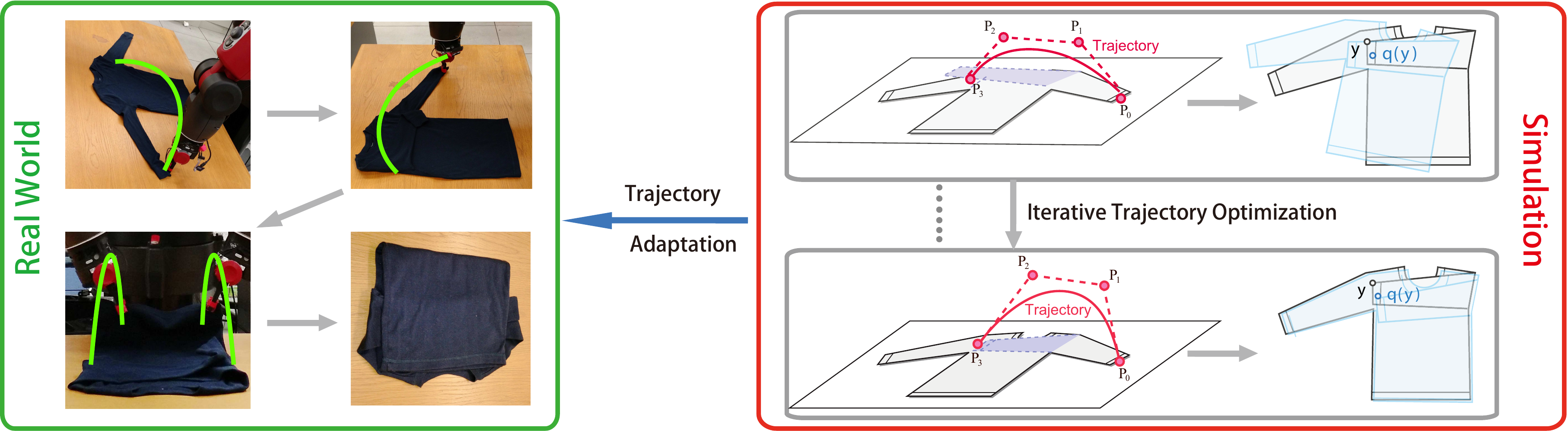}
  \caption{
		Details of the folding procedure. We apply offline simulation with iterative trajectory optimization to find the best trajectory for a specific folding action by comparing the result (light blue contour) with template (black contour). Similar steps are repeated until the garment is folded in the simulator. Then all the folding trajectories are exported, adapted, and implemented on a real robot. Green arcs illustrate the actual trajectories of robotic arms.}
\label{fig:folding_flowchart}
\end{figure*}

Figure~\ref{fig:folding_flowchart} shows the key steps of the garment folding.
The garment folding is the final step of the entire pipeline of garment manipulation which contains visual recognition, unfolding, ironing, and folding in Figure~\ref{fig:entire_pipeline}.
This section specifically addresses the robotic folding task (purple rectangle in Figure~\ref{fig:folding_flowchart}) with the goal of finding optimal trajectories to successfully fold garments.

Figure~\ref{fig:failure_examples} shows a few failure examples with improper trajectories.
We use green tape on the table to show the original position of the garments.
The first two rows show that if the moving trajectory is too low and close to the garment, the folded part will fall down, pull the rest, and cause drift of the whole garment. 
These cases usually happen when the folding step is lengthy without trajectory optimization.
The third row shows a case where the folding trajectory is too high, which will cause extra wrinkles or even piling up.
The last row shows two cases using two arms to fold.
If the arms are close to each other, the part in between loses tension, and will fall down and pull the rest away.
The focus of this work is to create trajectories for folding that will overcome these problems.


\subsection{Simulation Environment}
\label{section:Simulation}

\subsubsection{Folding Pipeline in Simulation}
In the model simulation, we use a physics engine~\cite{urlMaya} to simulate the movement and deformation of the garment mesh models.
We assume there is only one garment for each folding task, which has been placed flat on a table.
A virtual table is added to the scene which the garment lies on, as shown in Figure~\ref{fig:intro}, top.

During each folding step, the robot arm picks up a small part of the mesh, moves it to the target position following a computed trajectory, and places it on the table to simulate an entire folding scenario.
If the part of the garment to be folded is relatively wide, then both left and right arms may be involved.
The trajectory is generated using a B\'ezier curve, which will be discussed in section \ref{sec:trajectory_optimization} below.

\begin{figure}[!htpb]
\begin{center}
\includegraphics[width=0.95\linewidth]{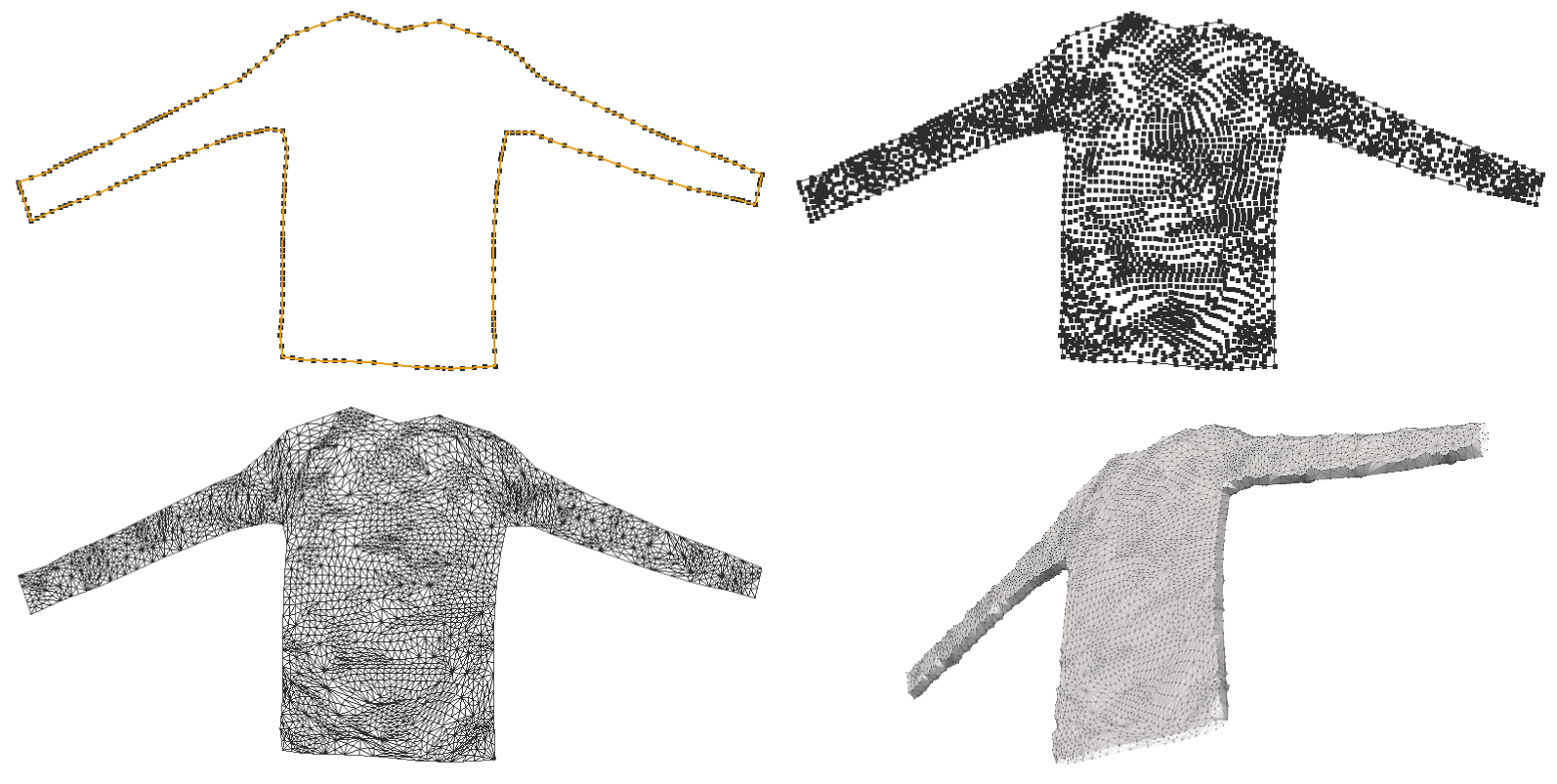}
\end{center}
   \caption{Garment models. {\sc{Top left:}} the input contour. 
   {\sc{Top right:}} we insert vertices into the internal region.
   {\sc{Bottom left:}} we build a flat triangle mesh using the contour and the inserted vertices.
   {\sc{Bottom right:}} we shift the contour vertices and mirror the mesh to create the garment mesh. 
}
\label{fig:garment_model}
\end{figure}

We can use the mesh model from the database to simulate the folding process.
However, for faster computation, these mesh models are relatively low resolution meshes, which are not very accurate when used to simulate folding via bending the mesh.
For more accurate simulation purposes, we propose a method to build a mesh model from our real garments.
Specifically, a garment mesh is created by first extracting the contour of the real garment~\cite{LiIROS2015}.
Then by inserting points on the inside of the garment contour, we triangulate a mesh by connecting these points.
Lastly, we mirror the mesh to construct a two-sided garment mesh (see figure~\ref{fig:garment_model}).

\subsubsection{Parameter Adaptation}

There are two key parameters needed to accurately simulate the real world folding environment. 
The first is the material properties of the fabric, and the second is the frictional forces between the garment and the table.

\paragraph{Material properties}
Through many experiments, we found that the most important property for the garments in the simulation environment is shear resistance.
It specifies the amount the simulated mesh model resists shear under strain; when the garment is picked up and hung by gravity, the total length will be elongated due to the balance between gravity force and shear resistance.
An appropriate shear resistance measure allows the simulated mesh to reproduce the same elongation as the real garment.
This measurement will bridge the gap between the simulation and the real world for the garment mesh model.

\begin{figure}[!htpb]
\begin{center}
 \includegraphics[width=0.45\textwidth]{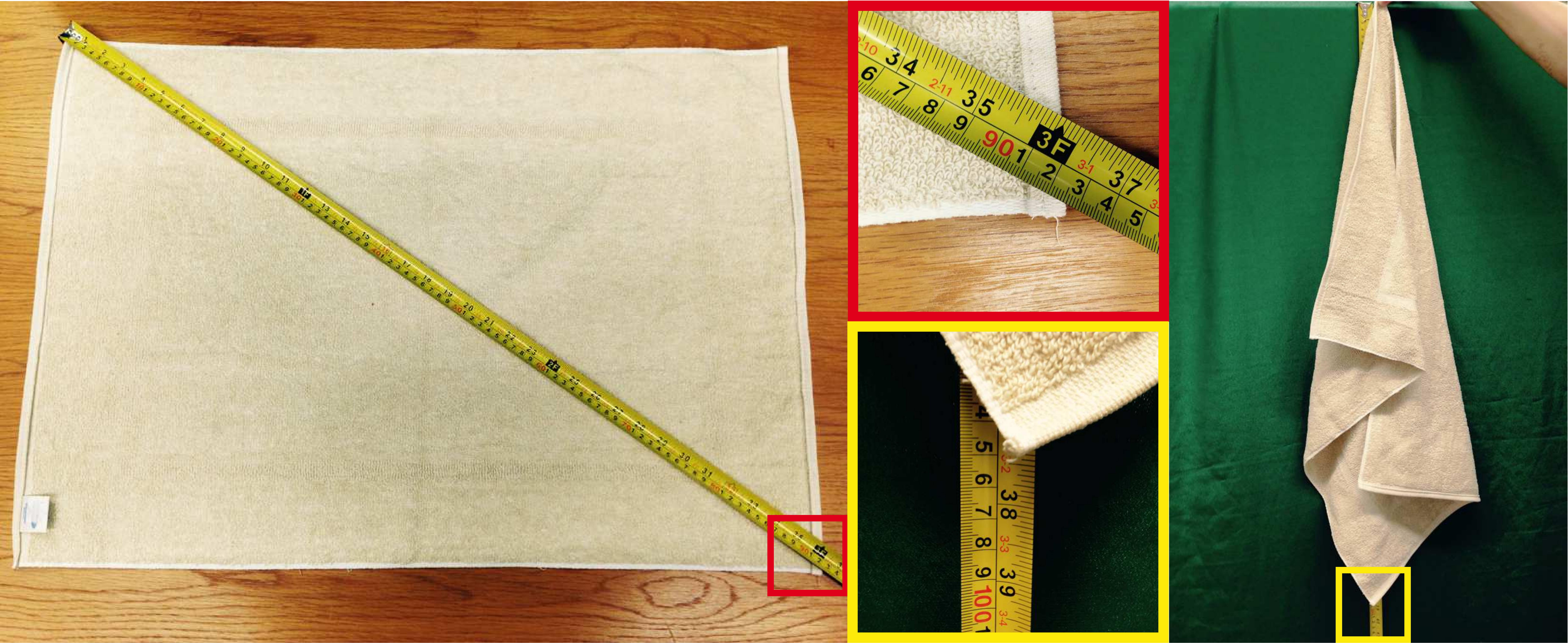}
\end{center}
   \caption{Method for measuring the shear resistance.
	{\sc{Left:}} Diagonal length measurement.
	{\sc{Middle:}} Zoomed in regions.
	{\sc{Right:}} The garment is hanging under gravity.}
\label{fig:measurement}
\end{figure}

For each new garment, we follow the steps described below to measure the shear resistance. 
Figure~\ref{fig:measurement} shows an example.
\begin{itemize}
\item[-] Manually pick one extremum part of the garment such as the sleeve end of a T-shirt, the waist part of a pair of pants, and a corner of a towel. 
\item[-] Hang the garment under gravity and measure the length between the picking point and the lowest point as $L_1$
\item[-] Slowly put down the garment on a table and keep the picking point and the lowest point in the previous step at maximum spread condition. Measure the distance between these two points again as $L_2$. The shear resistance fraction is defined as the following
\begin{align}
shear\_frac = ({L_1} - {L_2})/{L_2}
\end{align}
\item[-] We then the virtual garment into the same configuration in \emph{Maya}, adjusting the \emph{Maya} shear parameter such that the shear fraction as calculated in the simulator is identical to the real world. 
\end{itemize}

\paragraph{Frictional forces}
The surface of the table can be rough if covered by a cloth sheet or slippery if not covered, which leads to variance in friction between the table and garment.
A shift of the garment during the folding can possibly impair the whole process and cause additional repositioning.
Adjusting the frictional level in the simulation environment to the real world is crucial and necessary for trajectory optimization.

To measure the friction between the table and the garment, we do the following steps.
\begin{itemize}

\item[-] Place a real garment on the real table of length $L_t$.
\item[-] Slowly lift up one side of the real table, until the garment in the real world begins to slide. The lifted height is $H_s$. The friction angle is computed as,
\begin{align}
{\angle _{Friction}} = {\sin ^{ - 1}}({H_s}/{L_t})
\end{align}
\item[-] In the virtual environment, the garment is placed flat on a table with gravity. Assign a relatively high friction value to the virtual table. Lift up one side of the virtual table to the angle of ${\angle _{Friction}}$.
\item[-] Gradually decrease the frictional force in the virtual environment, until the garment begins to slide. 
Use this frictional force in the virtual environment as it mirrors the real world

\end{itemize}

With these two parameters set up, we obtain similar manipulation results for both the simulation and the real garment.

\subsection{Trajectory Optimization}
\label{sec:trajectory_optimization}

The goal of the folding task is specified by the initial and folded
shapes of the garment, and by the starting and target positions of the
grasp point (as in Figure~\ref{fig:traj_opt}). Given the simulation 
parameters, we seek the trajectory that effects the desired set of folds. 
We first describe how to optimize the trajectory for a single end 
effector and then discuss the case of two end effectors.

\subsubsection{Trajectory parametrization}
We use a B\'ezier curve~\cite{Farin:1988} to describe the trajectory.
An $n$-th order B\'ezier curve $\mathbf{T}(u)$ has $(n+1)$ control points 
$\mathbf{P}_k = ({P}_{k,x}, {P}_{k,y}, {P}_{k,z})^T\in \mathbb{R}^3$, defined by
\begin{align}
\mathbf{T}(u) = \sum_{k=0}^n B_k^n(u) \mathbf{P}_k,
\end{align}
where $B_k^n(u) = \begin{pmatrix} n \\ k \end{pmatrix} (1-u)^{n-k} u^k$ are the Bernstein basis
functions.

\begin{figure}[!htpb]
\centering
\includegraphics[width=0.95\linewidth]{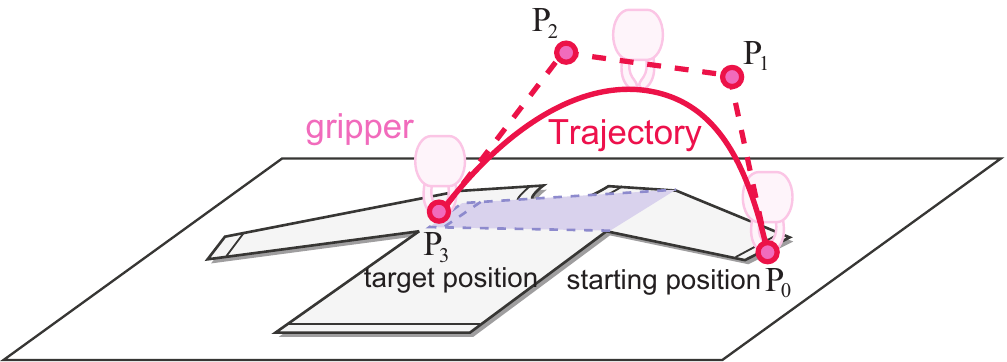}
\caption{An example of the folding task: we want to fold a sleeve into the blue
target position, by using a robotic gripper to move the tip of the sleeve 
(grasp point) from the starting position ($\mathbf{P}_0$) to the target position
($\mathbf{P}_3$), following a trajectory, shown as the red curve.
$\mathbf{P}_1$ and $\mathbf{P}_2$ are knot points that form the B\'ezier trapezoid.}
\label{fig:traj_opt}
\end{figure}

We use $n=3$ for simplicity, but our method can be easily extended to
deal with higher order curves. $\mathbf{P}_0$ and $\mathbf{P}_3$ are
fixed to the specified starting and target positions of the grasp
point (as in Figure~\ref{fig:traj_opt}). The intermediate control points $\mathbf{x} =
(\mathbf{P}_1^T, \mathbf{P}_2^T)^T$ can then be adjusted to define a new trajectory using the objective function defined below. 
\begin{align}
\label{eq:objective}
\mathbf{x}_{opt} = \argmin_{\mathbf{x}} \{\underbrace{ l_\mathbf{x} + \alpha D(\mathcal{S}_t, \mathcal{S}_{\mathbf{x}}) }_{C(\mathbf{x})} \}^2.
\end{align}


Here $C(\mathbf{x})$ is a cost function with two terms. The first term
penalizes the trajectory length 
$l_\mathbf{x}$, thus preferring a folding path that is efficient in
time and energy. The second term seeks the desired fold, by penalizing dissimilarity $D(\mathcal{S}_t, \mathcal{S}_{\mathbf{x}})$
between the desired folded shape $\mathcal{S}_t$, compared to
the shape $\mathcal{S}_{\mathbf{x}}$ obtained by the
candidate folding trajectory $\mathbf{x}$, as predicted by a cloth simulation; we used a physical simulation engine~\cite{urlMaya}, 
for the cloth simulation.
The weight $\alpha$ balances the two terms; we used $\alpha 
= 10^3$ in our experiment.

\begin{figure}[!htpb]
\centering
\begin{tabular}{cc}
\includegraphics[width=0.5\linewidth]{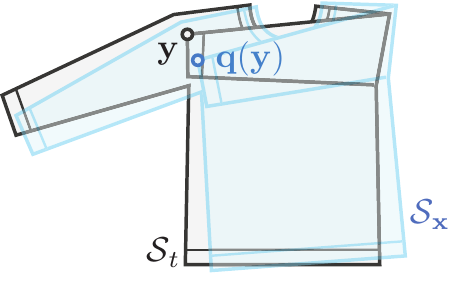} &
\includegraphics[width=0.4\linewidth]{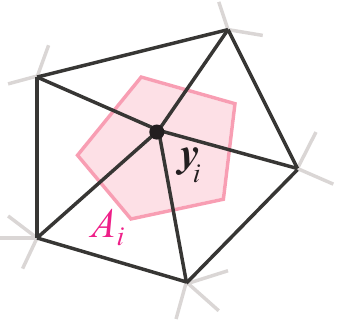} \\
\end{tabular}
\caption{
{\sc{Left: }}
The dissimilarity captures the misalignment between 
$\mathcal{S}_t$ and $\mathcal{S}_{\mathbf{x}}$ by integrating the distance 
between the corresponding points $\mathbf{y} \in \mathcal{S}_t$ and 
$\mathbf{q}(\mathbf{y}) \in \mathcal{S}_{\mathbf{x}}$ over the garment.
{\sc{Right: }}The barycentric dual area $A_i$ associated with this 
vertex $\mathbf{y}_i$ is defined as the area of the polygon created by
connecting the barycenters of the triangles adjacent to $\mathbf{y}_i$.}
\label{fig:dissimilarity_bary_dual}
\end{figure}

Intuitively, dissimilarity measures the difference between the desired folded shape and the folded garment in simulation.
We define the dissimilarity term as
\begin{align}\label{eq:dissimilarity-cont}
D(\mathcal{S}_t, \mathcal{S}_{\mathbf{x}}) = 
\frac{1}{|\mathcal{S}_t|}\int_{\mathcal{S}_t} \|\mathbf{q}(\mathbf{y}) - 
\mathbf{y}\| dA,
\end{align}
where $|\mathcal{S}_t|$ is the total surface area of the garment mesh including both sides of the garment, $\mathbf{y} \in
\mathcal{S}_t$ is a point on the target folded shape $\mathcal{S}_t$,
$\mathbf{q}(\mathbf{y}) \in \mathcal{S}_{\mathbf{x}}$ is the corresponding
point on the simulated folded shape, and $dA$ is the area measure, see 
Figure~\ref{fig:dissimilarity_bary_dual}, left. 
Our implementation assumes $\mathcal{S}_t$ 
and $\mathcal{S}_{\mathbf{x}}$ are given as triangle meshes, and 
discretizes~\eqref{eq:dissimilarity-cont} as
\begin{align}
\tilde{D}(\mathcal{S}_t, \mathcal{S}_{\mathbf{x}}) = 
\frac{1}{|\mathcal{S}_t|}\sum_i \|\mathbf{q}_i - \mathbf{y}_i\| A_i,
\end{align}
where $\mathbf{y}_i$ is the barycenter of $i$-th triangle on the target shape, 
$\mathbf{q}_i$ is the (corresponding) barycenter of $i$-th triangle on the simulated
shape, and $A_i$ is the area of the $i$-th triangle on the target shape.

To compute the trajectory length $l_\mathbf{x}$, we use the De Casteljau's
algorithm~\cite{Farin:1988} to recursively subdivide the B\'ezier curve 
$\mathbf{T}$ into a set of B\'ezier curves $\mathbf{T}^{(j)}$, until the 
deviation between the chord length ($\|\mathbf{P}_0^{(j)} - 
\mathbf{P}_3^{(j)}\|$) and the total length between the control points 
($\sum_{i=0}^2 \|\mathbf{P}_i^{(j)} - \mathbf{P}_{i+1}^{(j)}\|$) for each 
subdivided curve $\mathbf{T}^{(j)}$ is sufficiently small. Then, $l_\mathbf{x}$
is approximated by summing up the chord lengths of all the subdivided curves: 
$l_\mathbf{x} \approx \sum_j \|\mathbf{P}_0^{(j)} - \mathbf{P}_3^{(j)}\|$.

We initialize $\mathbf{P}_1$ and $\mathbf{P}_2$ as
\begin{align}
\mathbf{P}_1 = \frac{2}{3}\mathbf{P}_0 + \frac{1}{3}\mathbf{P}_3 + h \|\mathbf{P}_0-\mathbf{P}_3\| \mathbf{e}_v,\\
\mathbf{P}_2 = \frac{1}{3}\mathbf{P}_0 + \frac{2}{3}\mathbf{P}_3 + h \|\mathbf{P}_0-\mathbf{P}_3\| \mathbf{e}_v,
\end{align}
where $\mathbf{e}_v$ is the unit vector in the upward vertical direction,
$h$ is a constant, which is set to $1/3$, which means the initial trajectory will have equal horizontal extent between knot points.

\subsection{Optimization.}
\label{sec:optimization}
To optimize equation \eqref{eq:objective}, we apply a secant version of
the Levenberg-Marquardt algorithm~\cite{Madsen:2004}\cite{Nocedal:2006}. 
For the current trajectory generated by $\mathbf{x}$, we estimate the derivative $\nabla C(\mathbf{x})$ of
the cost function $C(\mathbf{x})$ numerically, by sampling slightly modified
trajectories $\mathbf{x} + \delta \mathbf{e}_j$, where $\mathbf{e}_j, 
1\leq j \leq \textrm{dim}(\mathbf{x})$, are the orthonormal bases, and we used 
$\delta = 10^{-1}$ in our implementation.

The secant version of Levenberg-Marquardt algorithm iteratively builds a local 
quadratic approximation of $\{C(\mathbf{x})\}^2$ based on the numerical
derivative, and then takes a step toward an improved state. The direction of
the step is a combination of the steepest gradient descent direction and the 
conjugate gradient direction. We use the specific approach described
by Madsen et al.~\cite{Madsen:2004} (see \S3.5 therein).
The iterative procedure terminates when the improvement in $\{C(\mathbf{x})\}^2$ becomes
sufficiently small.

In the case of using multiple arms, we associate an individual trajectory 
$\mathbf{x}_i$ to each of the arms $R_i$. We then extend the state variable
to $\mathbf{x} = (\mathbf{x}_1^T, ...)^T$. The rest of the optimization
procedure is the same as the single arm case. 
Note that both single and dual-arm trajectories are in 3D space.
The optimization for dual-arm trajectories is able to find a solution which will overcome failures such as shown in Figure~\ref{fig:failure_examples} bottom.


\subsection{Experimental Results}
\label{section:experiments}

To evaluate our results, we tested our method on several different garments such as long-sleeve t-shirts, pants, and towels for multiple trials, as shown in Figure~\ref{fig:garment_stat_fold} left.
These garments require both single and dual-arm folds.


\subsubsection{Measurement of parameters}
To make the offline simulation better approximate the real scenario, we manually measure the stretch resistance of each garment and friction on the table.

\begin{figure*}[t] 
  \begin{minipage}[]{0.4\textwidth} 
    \centering 
    \includegraphics[width=0.95\textwidth]{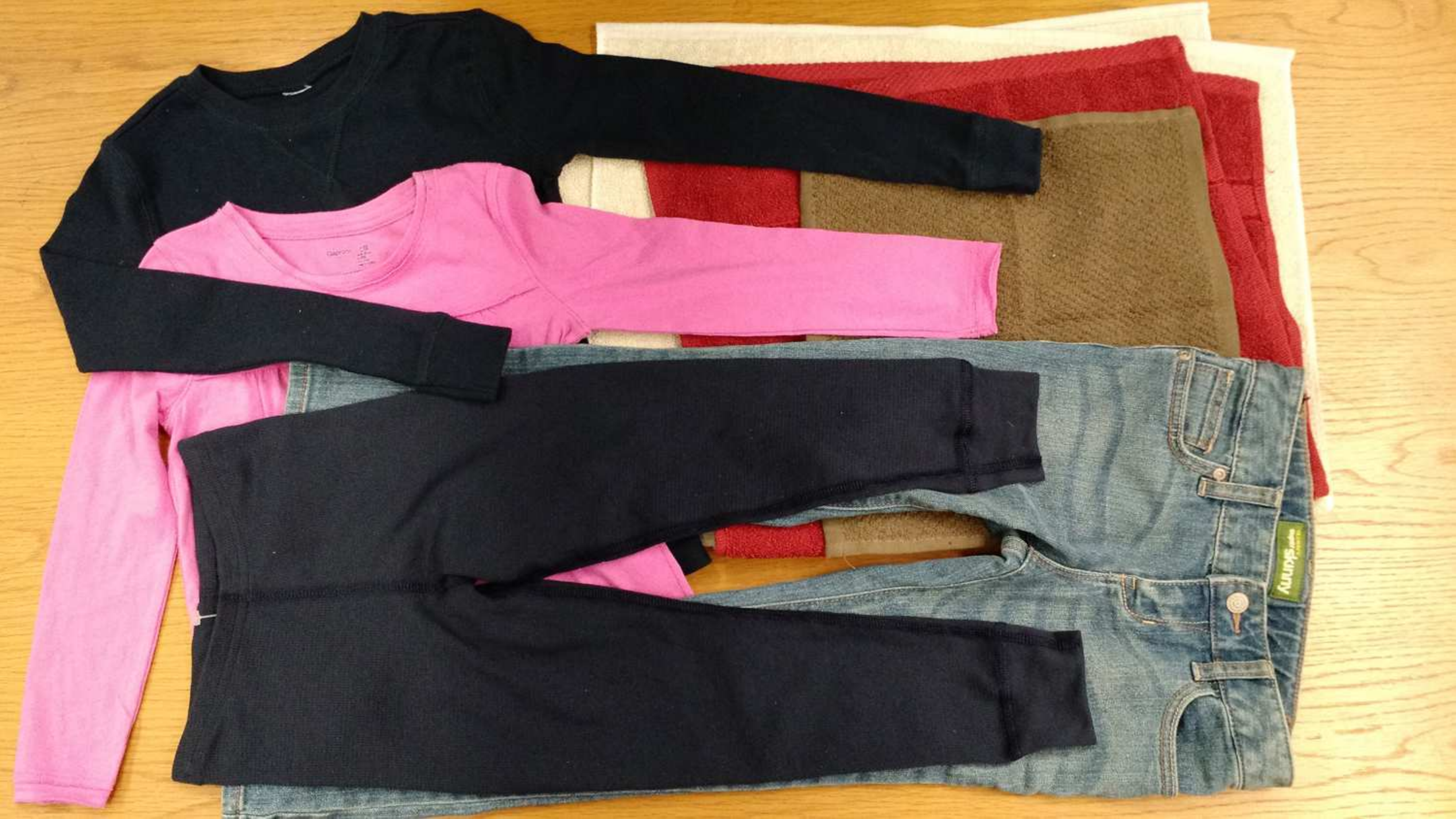} 
  \end{minipage}
  \begin{minipage}[]{0.55\textwidth} 
	\small
	\centering
    \begin{tabular}{c|p{0.8cm}|p{1.3cm}|p{1.6cm} | p{1.1cm}} \hline
      Garment Type &  \centering Stretch (\%) & \centering Friction Angle ($^{\circ}$) & \centering Maya Shear Resistance &\centering Maya Friction  \tabularnewline \hline
      Long-Sleeve T-Shirt (large) & \centering $2.9$  & \centering $24.3$  & \centering $200$ & \centering $0.7$ \tabularnewline
			Long-Sleeve T-Shirt (small) &  \centering  $2.9$ & \centering $24.7$ & \centering $200$ & \centering $0.7$ \tabularnewline
      Jeans &  \centering $2.9$  & \centering  $19.1$ & \centering $200$ & \centering $0.5$ \tabularnewline
			Pants &   \centering  $1.7$ & \centering  $21.9$ & \centering $340$ & \centering $0.6$ \tabularnewline
			Large Towel &  \centering $2.2$  & \centering  $18.7$ & \centering $260$ & \centering $0.5$ \tabularnewline
			Medium Towel &  \centering $3.1$  & \centering  $22.3$ & \centering $190$ & \centering $0.6$ \tabularnewline
			Small Towel &  \centering $1.1$  & \centering $24.3$  & \centering $530$ & \centering $0.7$ \tabularnewline \hline
			{\bf{Average}} & \centering ${\bf{2.4}}$ & \centering ${\bf{22.2}}$ & \centering $\bf{274}$ & \centering $\bf{0.6}$ \tabularnewline
      \hline
    \end{tabular}
  \end{minipage} 
	\caption{{\sc{Left:} }A picture of our test garments.
	{\sc{Right:}} Results for each unfolding test on the garments. 
	We show the results of stretch percentage, Friction angle of the table, and the corresponding parameters in Maya by each test.
	The last row shows the average of each measurement component.
	}
\label{fig:garment_stat_fold}
\end{figure*}

Figure~\ref{fig:garment_stat_fold}, left shows a picture of all the test garments we used in different colors, sizes, and material properties.
Figure~\ref{fig:garment_stat_fold}, right table shows the measured parameters of each test garment, including stretch percentage and Friction angle, and corresponding \emph{Maya} parameters.
For common garments, these parameters do not have a significant variance.
Therefore, we suggest that if researchers use simulators such as \emph{Maya}, the average values of each column are a reasonably good initialization.

\subsubsection{Garment manipulation and folding}

\begin{figure*}[!htbp]
\begin{center}
 \includegraphics[width=0.98\textwidth]{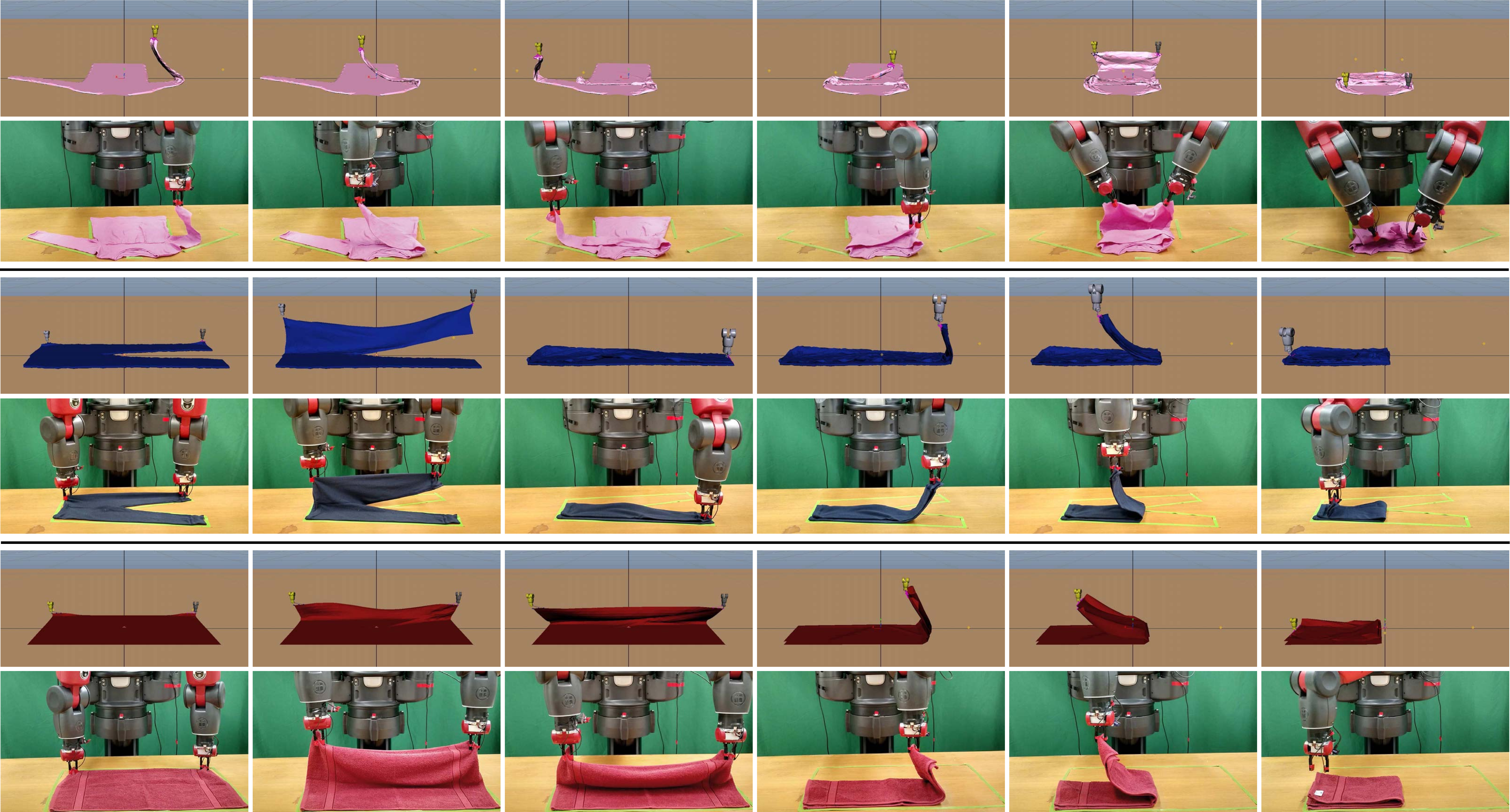}
\end{center}
   \caption{Successful folding examples with optimized folding trajectories from offline simulation. 
	The first row of each group is from the simulation and the second row is from the real world (Green tape shows the original garment contour position).
	{\sc{Top Group:}} Long-sleeve shirt folding with $3$ steps.
	{\sc{Middle Group:}} Long pants folding with $2$ steps.
	{\sc{Bottom Group:}} Medium size towel folding with $2$ steps.}
\label{fig:good_examples}
\end{figure*}

Figure~\ref{fig:good_examples} shows three successful folding examples from the simulation and the real world, including a long-sleeve shirt, a pair of pants, and a medium size towel. 
We show six key frames for each folding task.
The folding poses from the simulation are in the first row of each group with an optimized trajectory.
We also show corresponding results from the real world.
The green tape contour on the table indicates the original position of the garment.

\begin{figure}[t]
\begin{center}
 \includegraphics[width=0.48\textwidth]{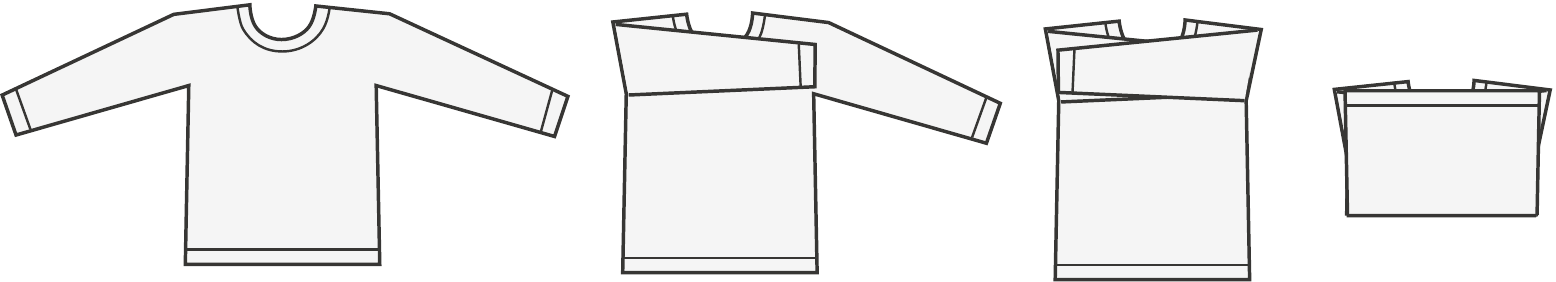}
\end{center}
   \caption{Garment folding plan for a long-sleeve T-shirt.}
\label{fig:folding_plan}
\end{figure}

Each garment is first segmented from the background and key points are detected from the binary mask.
Given the key points, a corresponding multi-step folding plan is created (The folding plan is predefined, and one of our folding plans for a long-sleeve T-shirt is shown in Figure~\ref{fig:folding_plan}).
For each garment, we have optimized trajectories for each folding step.
Here, we map these optimized trajectories to our scenario according to the generated folding plan.
Then the Baxter robot follows the folding plan with optimized trajectories to fold the garment.
We can see that the deformation of the real garment and the simulated garment is very similar.
Therefore, the final folding outcome is comparable to the simulation.

Table~\ref{fig:garment_experiments_stat} shows statistical results of the garment folding test.
Each time one or two robotic arms fold the garment counts as one fold.
We ran $10$ trials for each test garment.
It turns out that the folding performance of the Long-Sleeve T-Shirts and Towels are very stable with our optimized trajectories.
Jeans and pants are less stable because the shear resistance of the surface is relatively high, and sometimes is difficult to bend, leading to unsuccessful folding.
In the successful folding cases for jeans and pants, we sometimes ended up with small wrinkles, but the folding plan was still able to complete successfully.
We also show the average time to fold a garment in the last row. 
The robot is able to fold most garments in about $1.5$ minutes.

\subsubsection{Solution Space}
\label{sec:exp_solution_space}

The solution space is a subspace of the trajectory space where the folded garment ends in a shape with a dissimilarity score less than a threshold.
Intuitively, a number of trajectories within the solution space will fold the garment, leaving its shape close to the desired shape.
We have found that trajectories within the solution space can vary to a degree while still allowing the robot to accomplish the folding task. 
This result also agrees with the fact that people do not have to follow a unique trajectory to fold the garment.
However, trajectories outside the solution space cause issues for the folding task (see Figure~\ref{fig:failure_examples}). 
Our trajectory optimization automatically avoids such cases.

\begin{figure}[!htpb]
\centering
 \includegraphics[width=\linewidth]{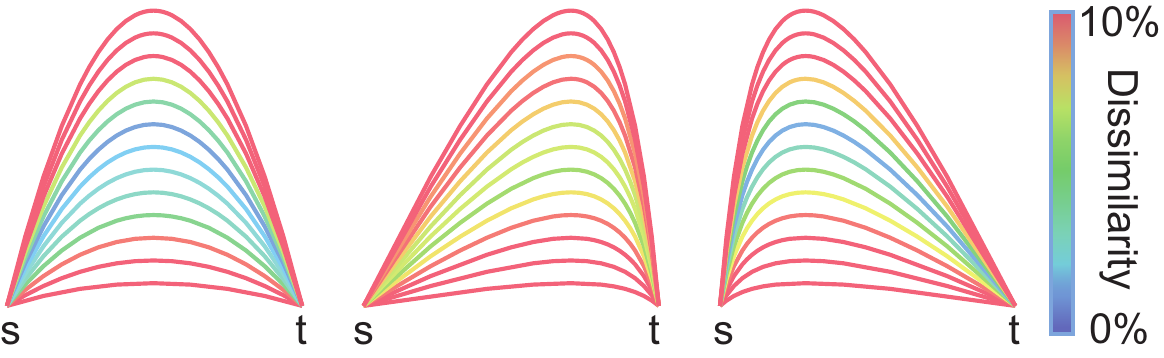}
 \caption{The dissimilarity values from different trajectories for folding the towel model in the second folding step. 
	The trajectory is projected to a 2D plane for illustration purposes.
	S and T stand for the start and target position, respectively. (Best viewed in color)}
\label{fig:solution_space}
\end{figure}

To further explore the relationship between the trajectories and folded shapes, we experimented the folding with a few different trajectories in simulation.
A notable finding is that the symmetric trajectories can always produce better folded shape, as shown in Figure~\ref{fig:solution_space}.
The thirteen color curves in each plot represent thirteen different trajectories. 
The dissimilarity bar on the right shows the difference between folded shape and the desired folded shape for each folding simulation. 
We also tested with asymmetric trajectories for the folding, as shown in the second and third plots in Figure~\ref{fig:solution_space}. 
We can see that the second plot has larger dissimilarities than the first and third plots, which is mainly caused by the friction. The robot should raise the starting point to a high enough position at the beginning to prevent the grasped portion of the garment pushing the other portion on the table.
This is also consistent with our simulation results that our optimizer will drive the height of the trajectories to a reasonable distance from the garment.

There is a trade-off between doing contour fitting at each step and total time spent to fold a garment.
In this work, we start with one template and then assume that each step after that the folded garment is close to that in the simulation.
Our experimental results as shown in Table~\ref{fig:garment_experiments_stat} verify that this method works well and is able to save time since we only do the contour fitting once.
With our simulated trajectories, the Baxter robot is able to fold a garment under predefined steps correctly.
An alternative method could use the contour fitting at each step but this would require more time and computation.

We note that some failures due to the motor control error from the Baxter robot.
When the robot executes an optimized trajectory, its arm suffers from a sudden drop or jitter.
Such actions will raise pull forces to the garment, leading to drift and inaccurate folding.
This can be solved by using an industrial level robotic arm with more accurate control.
We also note that failures can be recognized with the correct sensing suite, and we are currently investigating ways to effect online error recovery for such failures.
One difference between the simulation and the real world we found is that moving a point on the mesh in the simulation is different from using a gripper to grasp a small area of a real garment and move it.
In the future, we hope to be able to simulate a similar grasp effect for the trajectory optimization.

\begin{table}[t] 
	\centering
    \begin{tabular}{c|p{1cm}|p{1.5cm}|p{1.5cm} } \hline
      Garment Type &  \centering \# of folds & \centering Success Rate & \centering Avg. Time (sec) \tabularnewline \hline
      L-S T-Shirt (large) & \centering $3$  & \centering  $10/10$ & \centering  $121$  \tabularnewline
			L-S T-Shirt (small) &  \centering  $3$ & \centering $10/10$ & \centering  $118$  \tabularnewline
      Jeans &  \centering $2$  & \centering $7/10$  & \centering  $88$  \tabularnewline
			Pants &   \centering  $2$ & \centering  $8/10$ & \centering $88$ \tabularnewline
			Large Towel &  \centering $2$  & \centering  $10/10$ & \centering  $90$   \tabularnewline
			Medium Towel &  \centering $2$  & \centering  $10/10$ & \centering  $88$  \tabularnewline
			Small Towel &  \centering  $2$ & \centering  $10/10$ & \centering  $83$  \tabularnewline \hline
			{\bf{Average}} & \centering ${\bf{2.3}}$  & \centering ${\bf{9.3/10}}$ & \centering ${\bf{97}}$  \tabularnewline
      \hline
    \end{tabular}
	\caption{Results of folding test for each garment . We show the number of folding steps, successful rate, and total time of each garment. Each garment has been tested $10$ times. L-S stands for Long-Sleeve. The time is the average over all successful trials for each garment.
	}
\label{fig:garment_experiments_stat}
\end{table}

\section{Conclusion}
In this paper, we introduced a simulation database of common deformable garments to facilitate recognition and manipulation.
The database contains five different garments within three categories: sweater, pants, and shorts.
Each garment is fully simulated with a number of depth images and 3D mesh models for all the semantic labeled grasping points.
We demonstrated three applications of using the database to improve the recognition and the manipulation of deformable objects.
The first is training from the simulated mesh models to recognizing an unknown object by 3D shape-based features.
The second is applying the simulated mesh model to guide the iterative regrasping of the garment using both  rigid and non-rigid registrations.
The third is importing the mesh model into the simulator and computing optimized trajectories for manipulation of the deformable objects.  Ee extensively tested the three applications with designed experiments such as garment recognition via pick up, unfolding the garment to a known desired state and laying flat, and using pe-computed folding plans to fold it using a novel trajecotry optimization method that prevents common folding errors.
We have addressed all the phases of the pipeline in Figure \ref{fig:entire_pipeline} individually.  However, there are still some system and hardware issues that prevent the system from being a completely seamless pipeline.  These are due to 1) kinematic constraints on the Baxter robot which limits its ability to work with larger garments on a normal size table, and 2) our need to manually mount the iron on the robot hand for the ironing task.  

While the focus of our work has been on clothing, we want to underline the point that model-driven, feed forward prediction can work well in complex environments with many unknown states.  While we have not yet attempted this, we believe that the ideas in this paper can be ported to similar domains such as food handling (``soft deformable objects") and articulated rigid objects that have  multiple kinematic states.

We'd like to thank J. Weisz, J. Varley, and R. Ying for many discussions, P. M. Lopez for the help of folding plan.
We'd also like to thank NVidia Corporation, and Intel Corporation for the hardware support. 
This material is based upon work supported by the National Science
Foundation under Grant No. 1217904.

\bibliographystyle{plain}

\end{document}